\documentclass[10pt]{article}
\usepackage[letterpaper,margin=1in]{geometry}
\usepackage[utf8]{inputenc}
\usepackage[T1]{fontenc}
\usepackage{lmodern}
\usepackage{microtype}
\usepackage{amsmath}
\usepackage{amssymb}
\usepackage{graphicx}
\usepackage{xcolor}
\usepackage{booktabs}
\usepackage{longtable}
\usepackage{array}
\usepackage{adjustbox}
\usepackage{enumitem}
\usepackage{listings}
\usepackage[authoryear,round]{natbib}
\usepackage{caption}
\usepackage[colorlinks=true,linkcolor=blue!55!black,citecolor=blue!55!black,urlcolor=blue!55!black,bookmarksnumbered=true]{hyperref}

\newcommand{\mdw}[1]{\dimexpr #1\linewidth-2\tabcolsep\relax}
\setlist[itemize]{leftmargin=1.6em,itemsep=0.15em,topsep=0.3em,parsep=0.1em}
\setlist[enumerate]{leftmargin=1.9em,itemsep=0.15em,topsep=0.3em,parsep=0.1em}
\definecolor{mdrulegray}{gray}{0}
\definecolor{mdarrowgray}{gray}{0}
\newif\ifshowtodos
\showtodostrue
\DeclareRobustCommand{\todo}[2][TODO]{\ifshowtodos\begingroup\def\_{\textunderscore\allowbreak{}}%
  \textcolor{red}{[#1\if\relax\detokenize{#2}\relax\else: #2\fi]}\endgroup\fi}

\newcounter{mdbox}
\renewcommand{\themdbox}{\arabic{mdbox}}
\newsavebox{\mdboxbox}

\makeatletter
\newif\if@mdappendix
\@mdappendixfalse
\let\mdorig@appendix\appendix
\renewcommand{\appendix}{\mdorig@appendix\@mdappendixtrue}
\newcommand{\mdsec@section}{}
\renewcommand{\@seccntformat}[1]{%
  \if@mdappendix\ifcsname mdsec@#1\endcsname Appendix~\fi\fi
  \csname the#1\endcsname
  \if@mdappendix\ifcsname mdsec@#1\endcsname .\fi\fi
  \quad}
\makeatother
\newif\ifmdbibprinted
\mdbibprintedfalse
\newif\ifmdcitations
\mdcitationsfalse
\newcommand{\mdBibStyle}{unsrtnat}
\newcommand{\mdBibFiles}{refs,refs_report}
\newcommand{\mdBibliography}{\ifmdcitations\ifmdbibprinted\else\mdbibprintedtrue
  \bibliographystyle{\mdBibStyle}\bibliography{\mdBibFiles}\fi\fi}

\usepackage{startlux-report}

\usepackage{authblk}
\title{StartLux-Decision: AutoThink Decision Models with Reinforcement Learning for Calibrated Deliberative Decisions}

\renewcommand{\Affilfont}{\normalsize}
\makeatletter
\def\AB@affilsepx{\quad\ \protect\Affilfont}
\makeatother
\author[1,2]{Xialie Zhuang}
\author[1,2]{Zunhai Su}
\author[2]{Jia Pan}
\author[2,3]{Ngai Wong}
\author[4,*]{Guanhua Chen}
\author[1,*]{Chuan-Wei Kuo}
\affil[1]{StartLux}
\affil[2]{The University of Hong Kong}
\affil[3]{EMUS Lab}
\affil[4]{Southern University of Science and Technology}
\date{}
\hypersetup{
pdftitle={StartLux-Decision: AutoThink Decision Models with Reinforcement Learning for Calibrated Deliberative Decisions},
pdfauthor={Xialie Zhuang, Zunhai Su, Jia Pan, Ngai Wong, Guanhua Chen, Chuan-Wei Kuo}
}
\mdcitationstrue

\begin{document}
\maketitle

\begingroup
\renewcommand{\thefootnote}{*}
\footnotetext{Corresponding authors.}
\endgroup

\begin{mdresources}
\mdresource{Models}{\href{https://huggingface.co/startlux-models}{huggingface.co/startlux-models}}
\mdresource{Code}{\href{https://github.com/StartLuxLabs/StartLux-Decision}{github.com/StartLuxLabs/StartLux-Decision}}
\mdresource{Blog}{\href{https://decision.startlux.com/}{decision.startlux.com}}
\end{mdresources}

\begin{abstract}
We introduce StartLux-Decision, state-of-the-art decision models spanning 0.8B--27B parameters, combining efficient single-pass prediction with adaptive, confidence-guided reasoning.
The \textbf{\textit{Instant}} models leverage bidirectional attention in full-attention layers to improve decision accuracy while preserving efficient single-pass inference.
The \textbf{\textit{AutoThink}} models selectively invoke explicit reasoning for uncertain decisions while answering confident cases directly.
We develop a three-stage training pipeline that establishes shared decision capabilities, specializes the two variants, and equips AutoThink with explicit reasoning through reasoning supervision and reinforcement learning.
To improve reasoning efficiency while preserving decision quality, we introduce Reinforcement Learning for Calibrated Deliberative Decisions (\rlcdtwo{}).
Extending decision-oriented reinforcement learning to explicit reasoning, \rlcdtwo{} uses a frozen model and proper scoring rules to reward improvements in decision probability quality, while imposing a soft penalty on overlong reasoning traces.
\rlcdtwo{} increases the proportion of reasoning traces that terminate within the generation limit by up to 48 percentage points and reduces average reasoning length by up to 65\%, while largely preserving decision accuracy.
On the public subset of Decision Index 0.3, the Instant models outperform the strongest evaluated comparable-scale open-weight decision models by 7.4--16.9 points, with the 27B AutoThink model further improving its score to 73.4.
We also introduce DecisionPivot, a programmatically verifiable benchmark of computational decisions featuring contrast pairs that differ in a single answer-changing detail.
On DecisionPivot, AutoThink consistently outperforms its Instant counterparts by 11.5--18.4 percentage points in pair accuracy and achieves leading performance among evaluated systems, highlighting the benefits of adaptive reasoning for computation-intensive decisions.
To facilitate practical deployment, we support inference across vLLM, llama.cpp, and MLX, together with quantized model variants for reduced memory requirements.
We envision StartLux-Decision advancing a new generation of decision models that adaptively allocate reasoning to challenging decisions, balancing decision quality, computational efficiency, and reliability.
\end{abstract}

\begin{center}
\vspace{-6mm}
\adjincludegraphics[max width=\linewidth,max height=0.47\textheight]{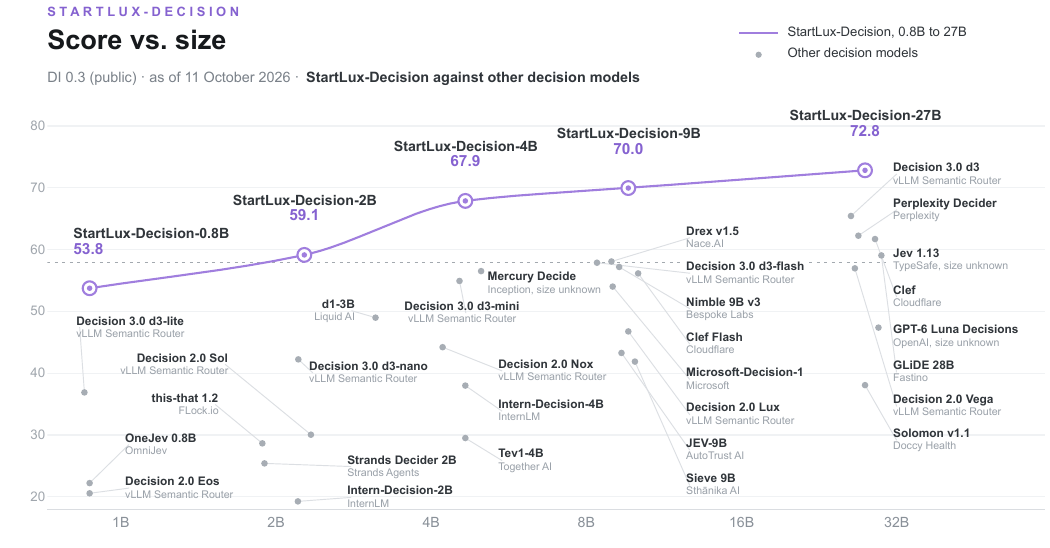}
\vspace{-7mm}
\captionof{figure}{\textbf{StartLux-Decision leads at every evaluated scale on the public subset of Decision Index 0.3.}}
\label{fig:decision-index-summary}
\end{center}

\clearpage
\tableofcontents
\clearpage

\section{Introduction}\label{sec:introduction}
\vspace{-3mm}
Decision-making is central to modern large language model (LLM) systems, underpinning tasks such as request routing, retrieval assessment, tool-call validation, and agent outcome verification~\citep{gu2024survey}.
A \emph{decision model} supports these operations by answering typed questions about an input state and producing probability distributions over candidate options~\citep{tang2026typed}.
Beyond identifying the most likely outcome, these probabilities enable downstream systems to decide whether to act autonomously, invoke a stronger model, or defer to human review.
Existing open-weight decision models largely rely on single-pass prediction without explicit reasoning~\citep{cloudflare2026clef,perplexity2026decider}.
While efficient, this approach can struggle with computation-intensive decisions involving arithmetic, scheduling, and procedural verification, where multi-step reasoning may be essential~\citep{wei2022chain,snell2024scaling}.
Moreover, we observe that decision-oriented training can substantially impair a model's ability to generate and reliably terminate reasoning traces.
Enabling token generation alone is insufficient to recover effective reasoning.
These challenges raise a central question: \emph{How can decision models incorporate reliable reasoning when needed while retaining efficient single-pass prediction and probabilistic decision outputs?}

\begin{figure}[t]
\centering
\adjincludegraphics[max width=1\linewidth,max height=0.5\textheight]{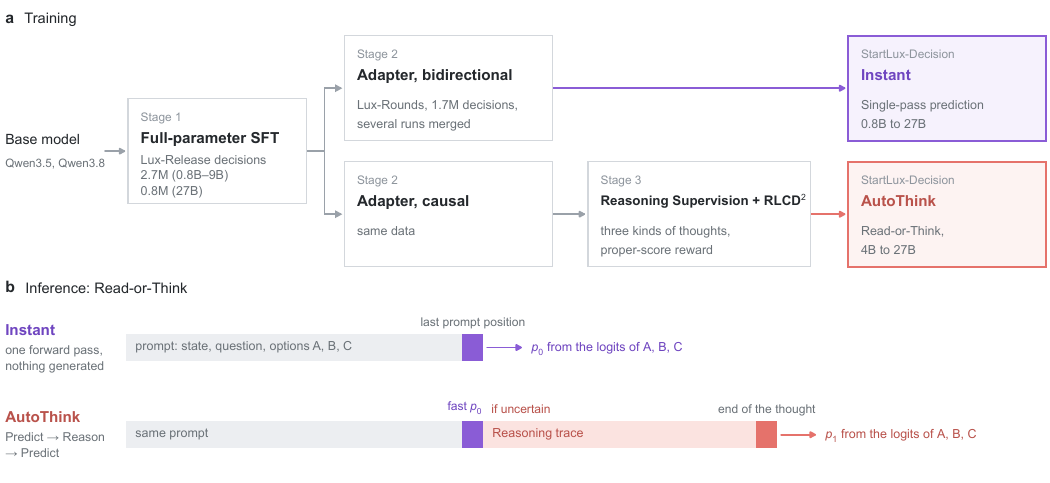}
\caption{\textbf{Training and inference workflows of StartLux-Decision.} (a) Three-stage training establishes shared decision capabilities, specializes the Instant and AutoThink variants, and develops explicit reasoning in AutoThink through reasoning supervision and reinforcement learning. Instant adopts bidirectional attention in full-attention layers, while AutoThink retains causal attention. (b) Instant produces decision probabilities in a single forward pass, whereas AutoThink first obtains an initial prediction $p_0$ and selectively invokes reasoning based on its confidence. When reasoning is invoked, the post-reasoning prediction $p_1$ is combined with $p_0$ to form the final decision.}

\label{fig:model-workflows}
\end{figure}

To address this challenge, we introduce StartLux-Decision, a family of decision models spanning 0.8B--27B parameters, with two complementary variants sharing a probability interface.
The \textbf{\textit{Instant}} models leverage bidirectional attention in full-attention layers to improve decision accuracy while preserving efficient single-pass inference.
This allows earlier prompt representations to incorporate information from subsequent questions and candidate options.
The \textbf{\textit{AutoThink}} models selectively invoke explicit reasoning for uncertain decisions while answering confident cases directly, combining initial and post-reasoning probabilities when needed.
A three-stage training pipeline establishes shared decision capabilities, specializes the two variants, and equips AutoThink with explicit reasoning through reasoning supervision and reinforcement learning (Figure~\ref{fig:model-workflows}; Section~\ref{sec:model-design}).

We introduce Reinforcement Learning for Calibrated Deliberative Decisions (\rlcdtwo{}), building on Reinforcement Learning for Calibrated Decisions (RLCD)~\citep{almeida2026introducing} to extend decision-oriented reinforcement learning to explicit reasoning.
Rather than rewarding answer correctness alone, \rlcdtwo{} optimizes the reasoning policy for the quality of the probabilistic decisions it supports.
A frozen copy of the starting model provides decision probability readouts before and after reasoning, with their proper-score difference defining the decision-quality reward.
A soft length penalty further discourages unnecessarily long reasoning traces, balancing decision quality with computational efficiency.
Empirically, \rlcdtwo{} increases the proportion of reasoning traces terminating within the generation limit by up to 48 percentage points and reduces average reasoning length by up to 65\%, while largely preserving decision accuracy (Section~\ref{sec:rlcd}).

StartLux-Decision demonstrates strong performance across established decision benchmarks.
On the public subset of Decision Index 0.3, the Instant models outperform the strongest evaluated comparable-scale open-weight decision models by 7.4--16.9 points, with the 4B Instant model surpassing all evaluated reference decision models (Figure~\ref{fig:decision-index-summary}).
Aggregate benchmark scores, however, can obscure sensitivity to small but decision-critical input changes.
We therefore introduce DecisionPivot~\citep{startlux2026decisionpivot}, a programmatically verifiable benchmark of computational decisions featuring contrast pairs that differ in a single answer-changing detail.
By requiring both decisions in each contrast pair to be correct, DecisionPivot evaluates whether models can reliably distinguish between closely related cases with different correct answers.
On DecisionPivot, AutoThink consistently outperforms its Instant counterparts by 11.5--18.4 percentage points in pair accuracy, with the 27B AutoThink model reaching 86.9\% and leading all evaluated reference systems (Section~\ref{sec:experimental-evaluation}).

Beyond predictive accuracy, the reliability of decision probabilities also depends on how downstream systems select and route predictions.
Calibration of individual predictions does not necessarily carry over to these downstream policies.
For example, predictions assigned 80--90\% confidence by our 9B verifier achieve 92\% accuracy when evaluated individually, but only 78\% after selecting the highest-scoring response from 32 candidates.
This discrepancy reveals how downstream selection can distort confidence--accuracy relationships even when the underlying model remains unchanged.
We introduce policy-aware calibration, which incorporates selection and routing information to calibrate confidence under the deployed policy, reducing confidence--accuracy gaps to within 2.7 percentage points across 184 settings spanning four decision pipelines (Section~\ref{sec:decision-reasoning-analysis}).
For practical deployment, we support both variants across vLLM, llama.cpp, and MLX.
We also provide quantized model variants to reduce memory requirements and facilitate deployment across diverse hardware environments (Section~\ref{sec:inference-deployment}).

Together, these contributions establish StartLux-Decision as a state-of-the-art decision model family that combines strong predictive performance with efficient single-pass inference and adaptive reasoning through a shared probabilistic decision interface.
More broadly, we envision a new generation of decision models that not only make accurate predictions but also determine when additional reasoning is warranted and how confidently their outputs should guide downstream actions, balancing decision quality, computational efficiency, and reliability.
We publicly release the model weights and inference code to enable independent evaluation, facilitate practical deployment, and advance further research on adaptive and reliable decision-making.
Our implementation and deployment resources are available at \url{https://github.com/StartLuxLabs/StartLux-Decision}.

\section{Model Design and Training}\label{sec:model-design}
\suppressfloats[t]
StartLux-Decision comprises two complementary variants, Instant and AutoThink, designed for efficient single-pass prediction and adaptive reasoning, respectively.
We develop a novel three-stage training pipeline that establishes foundational decision capabilities, specializes the two variants through task and attention adaptation, and equips AutoThink with explicit reasoning through reasoning supervision and reinforcement learning.

\subsection{Decision Formulation and Probability Readout}\label{sec:decision-readout}
Existing open-weight decision models typically answer structured questions about an input state by assigning probabilities to candidate options, often through a single forward pass~\citep{tang2026typed,cloudflare2026clef,perplexity2026decider}.

\paragraph{Typed decisions.}
A decision request comprises an input state and a set of named, typed questions concerning that state.
The input state may be represented as a string or any JSON value, optionally accompanied by images.
Each question specifies a choice among named options, a binary decision, or a level on an ordered scale with 2 to 10 levels.
Candidate options may include natural-language criteria that further define their intended semantics.
For each question, the model produces a probability distribution over candidate options, providing both a predicted outcome and a confidence estimate to guide downstream decisions.

\paragraph{One-pass readout.}
Existing open-weight decision models typically produce probability distributions over candidate options through a single forward pass, without autoregressive token generation.
These probabilities can be obtained through dedicated prediction heads or directly from the output logits of a pretrained language model.
In the latter approach, each question is formatted as a prompt containing the input state, the question, and candidate options mapped to output tokens, whose logits are read from the language-model head at the final prompt position.
For a question with $K$ candidate options, let $z_k$ denote the logit associated with option $k$.
The decision probabilities can then be obtained by applying temperature scaling and a softmax over the candidate options:
\begin{equation}
\label{eq:decision-probability}
p_k = \frac{\exp(z_k/T)}{\sum_{j=1}^{K}\exp(z_j/T)},
\qquad k=1,\ldots,K,
\end{equation}
where $T>0$ is a temperature parameter that can be fitted on held-out data for probability calibration.

\subsection{Model Variants and Adaptive Inference}\label{sec:model-variants}

\paragraph{Shared probability readout.}
StartLux-Decision adopts an option-letter probability readout that maps each decision question to a separate prompt containing the input state, question, and candidate options labeled with single letters (A--Z).
The corresponding logits are extracted at the final prompt position through the existing language-model output projection, without introducing an additional prediction head.
The same readout is used for Instant's single-pass predictions and AutoThink's initial and post-reasoning probability estimates.
Each readout supports up to 26 candidate options.
Longer lists are processed in parallel groups of at most 25 options, with the top three candidates from each group advancing to a final round (Appendix~\ref{sec:D.4}).
Questions within a request are processed as separate batch rows.
Both variants support contexts of up to 262,144 tokens and process images through the pretrained vision tower, which remains frozen during training.

\paragraph{Instant models.}
The \textbf{\textit{Instant}} models follow the single-pass decision paradigm, producing decision probabilities without autoregressive token generation.
Since the complete prompt is available before prediction, we enable bidirectional attention in the full-attention layers, allowing earlier prompt representations to incorporate information from subsequent questions and candidate options.
The linear-attention layers remain causal, preserving the underlying hybrid architecture while adapting full-attention computation to the single-pass decision setting.

\begin{figure}[!t]
\centering
\adjincludegraphics[max width=\linewidth,max height=0.5\textheight]{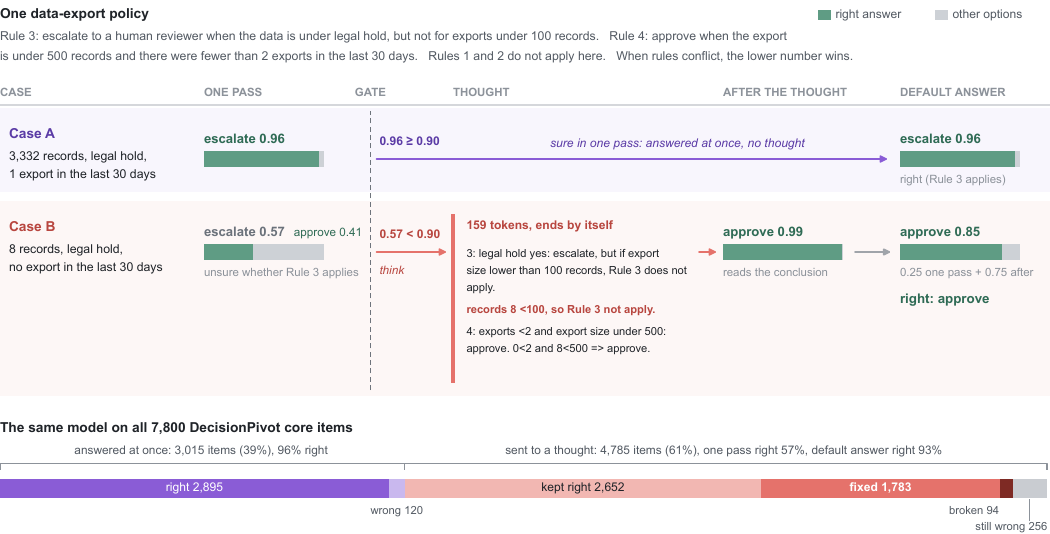}

\caption{\textbf{AutoThink selectively invokes reasoning for uncertain decisions.} Top: Two cases from the same DecisionPivot policy illustrate confidence-guided inference, with direct prediction for a confident case and reasoning-assisted correction for an uncertain case. Bars show decision probabilities, with correct options highlighted in green. Bottom: Across 7,800 core DecisionPivot items, the 27B AutoThink model makes direct predictions on 39\% of cases with 96\% accuracy and invokes reasoning for the remaining 61\%. On reasoning-invoked cases, accuracy improves from 57\% before reasoning to 93\% after default-answer fusion, with 1,783 initial errors corrected and 94 new errors introduced.}
\label{fig:read-or-think-example}
\end{figure}

\paragraph{AutoThink models and the default answer.}
The \textbf{\textit{AutoThink}} models extend single-pass prediction through \emph{Read-or-Think}, a confidence-guided inference mechanism that selectively invokes explicit reasoning for uncertain decisions while answering confident cases directly.
For each question, AutoThink first performs a causal one-pass readout to obtain an initial probability distribution $p_0$ over candidate options.
If the maximum predicted probability is at least the confidence threshold $g=0.9$, the model returns $p_0$ without additional reasoning.
Otherwise, it reprocesses the prompt with reasoning enabled, generates a reasoning trace of up to 4,096 tokens, and obtains a post-reasoning distribution $p_1$ from the option-letter logits following the trace.
The reasoning process and both probability readouts share the same model weights.
When reasoning is invoked, AutoThink combines the initial and post-reasoning probability distributions to form its final prediction, rather than relying exclusively on the post-reasoning estimate.
The final \emph{default answer} is defined as
\begin{equation}
\label{eq:autothink-default}
p_{\mathrm{default}} =
\begin{cases}
p_0, & \max_k p_{0,k}\geq g,\\
(1-\alpha)p_0+\alpha p_1, & \max_k p_{0,k}<g,
\end{cases}
\qquad g=0.9,\quad \alpha=0.75.
\end{equation}
The confidence gate is applied independently to each question, and the runtime closes any reasoning trace that reaches the generation limit without terminating.
Section~\ref{sec:adaptive-reasoning} examines the effects of the gate threshold and fusion weight.
Figure~\ref{fig:read-or-think-example} illustrates Read-or-Think using two cases from the same data-export policy in DecisionPivot.
The 27B AutoThink model answers the confident case directly but invokes reasoning for the uncertain case, where additional computation changes the predicted decision.
Across the 7,800 core DecisionPivot items, the model answers 39\% of questions directly and improves accuracy from 57\% to 93\% on the remaining reasoning-invoked cases.
Table~\ref{tab:reasoning-example} provides the corresponding probability readouts and generated reasoning trace.

\begin{table}[t]
\centering
\caption{\textbf{Selective reasoning corrects an initial decision error in policy interpretation.} A 27B AutoThink example from the programmatically generated DecisionPivot benchmark illustrates how reasoning identifies a policy exception and corrects an initially incorrect prediction. The reasoning trace is reproduced verbatim after the runtime's fixed initial line, without added line breaks. The final row provides a contrasting confident case under the same policy, answered without reasoning.}
\label{tab:reasoning-example}
{\footnotesize\setlength{\tabcolsep}{5pt}\renewcommand{\arraystretch}{1.1}
\noindent\begin{adjustbox}{max width=\linewidth}
\begin{tabular}{>{\raggedright\arraybackslash}p{\mdw{0.1474}}>{\raggedright\arraybackslash}p{\mdw{0.8383}}}
\toprule
\textbf{Step} & \textbf{What the model reads and writes} \\
\midrule
\mdbodyrow Policy & (1) When the verified requester flag is set to no and the number of exports in the last 30 days is 3 or more: Deny the request. (2) When the legal hold flag is set to yes and the data region is anything other than Branzim and the number of exports in the last 30 days is greater than 3: Deny the request. (3) When the case has the legal hold flag set to yes: Escalate the request to a human reviewer. (4) When the number of exports in the last 30 days is less than 2 and the export size is under 500 records: Approve the request. If the export size is lower than 100 records, Rule 3 does not apply. When rules conflict, the rule with the lower number wins. If no rule applies, the request is escalated to a human reviewer. \\
\mdbodyrow Uncertain case & 8 records, 0 exports in the last 30 days, data region Branzim, legal hold yes, requester verified no, and seven other fields that no rule mentions \\
\mdbodyrow Question & What does the policy require for this case? Options approve, deny and escalate. The correct answer is approve. \\
\mdbodyrow Single-pass readout & \textbf{escalate 0.567}, approve 0.408, deny 0.026. The maximum probability is below 0.9, so the model invokes reasoning. \\
\mdourrow \textbf{Reasoning trace, 159 tokens} & We need answer user's question. Need analyze policy. Need choose exactly one letter. Case: records 8, exports\_last\_30\_days 0, data\_region Branzim, legal\_hold yes, requester\_verified no. Rules: 1: verified no and exports \textgreater{}=3? no (0) 2: legal hold yes and region other than Branzim and exports \textgreater{}3? no 3: legal hold yes: escalate, but if export size lower than 100 records, Rule 3 does not apply. records 8 \textless{}100, so Rule 3 not apply. 4: exports \textless{}2 and export size under 500: approve. 0\textless{}2 and 8\textless{}500 =\textgreater{} approve. \\
\mdbodyrow Post-reasoning readout & \textbf{approve 0.995}, escalate 0.005, deny 0.001 \\
\mdbodyrow Default answer & \textbf{approve 0.848}, escalate 0.145, deny 0.007; the predicted option is correct \\
\mdbodyrow Confident case & Same policy; 3,332 records, 1 export in the last 30 days: \textbf{escalate 0.959} in one pass. The prediction is correct and exceeds the confidence threshold, so reasoning is not invoked \\
\bottomrule
\end{tabular}
\end{adjustbox}\par}
\end{table}

\subsection{Three-Stage Training Pipeline}\label{sec:model-training}
Our three-stage training pipeline develops the complementary capabilities of Instant and AutoThink.
Stages 1 and 2 establish and specialize probabilistic decision prediction, while Stage 3 equips AutoThink with explicit reasoning through supervision and reinforcement learning, preserving its single-pass prediction capability.

\subsubsection{Stage 1: Full-Parameter SFT}\label{sec:stage1-sft}
Stage 1 establishes the foundational decision capabilities shared by all StartLux-Decision models through full-parameter supervised fine-tuning (SFT) on Lux-Release, our initial training mixture.
\paragraph{Training Data.}
Lux-Release integrates diverse public datasets reformulated as typed decisions, spanning classification, natural language inference, question answering, safety, preference modeling, and tool-use tasks.
Additional sources include community collections of typed decisions, benchmark training splits, multiple-choice questions covering knowledge and commonsense reasoning, and a smaller portion of programmatically generated examples (Appendix~\ref{sec:A}).
The training mixture contains approximately 2.7 million examples for models ranging from 0.8B to 9B parameters, whereas the 27B model uses approximately 0.8 million examples, with a greater proportion of benchmark training splits and multiple-choice questions.
To mitigate answer-position bias, we deterministically permute candidate options for each example using a hash-based procedure.
To reduce training--evaluation overlap, we exclude examples matching the Decision Index test sets or any of the 231 public JevBench v1.4.2 items~\citep{benchmarkheaven2026jevbench}, together with all other examples belonging to the same parent groups (Appendix~\ref{sec:A}).
\paragraph{Training Procedure.}
We fine-tune all language-model parameters, including the output projection, using soft cross-entropy between the target and predicted probability distributions over candidate options at the final prompt position.
This objective directly optimizes the option-letter probability readout without introducing an additional prediction head.
Following training, we fit a separate temperature for each question type on held-out data to calibrate the predicted probabilities without changing the ranking of candidate options.
All fitted temperatures exceed one, indicating a consistent tendency toward overconfidence before calibration.

\subsubsection{Stage 2: Adapter-Based Decision Training}\label{sec:stage2-decision-training}
Stage 2 extends the shared decision foundation along two complementary dimensions: broader task coverage and attention adaptation.
We train adapters on Lux-Rounds, our second training mixture, and subsequently merge them into the Stage 1 model weights.
The Instant models adopt bidirectional attention in full-attention layers, while the AutoThink models retain causal attention to support subsequent reasoning training.

\paragraph{Training Data.}
Lux-Rounds comprises approximately 1.7 million decision examples designed to broaden task coverage beyond Stage 1.
Most examples are drawn from benchmark training splits and related public datasets, supplemented by \emph{breadth} data covering decision types not represented in the Decision Index.
Additional sources include programmatically generated procedural decisions and select-all questions, API-generated examples, and teacher-labeled decisions.
We retain API-generated examples only when two independent verifier models agree with their labels.
A subset of Lux-Release is also retained to preserve the capabilities established in Stage 1.
GSM8K contributes only training problems formatted according to the revised DI 0.3 benchmark.

\paragraph{Training Procedure.}
We optimize the adapters using the same decision objective as Stage 1 before merging them into the model weights.
Each released Instant model is obtained by uniformly averaging the weights of three independently trained runs at 0.8B or four runs at 2B--27B, differing only in their random seeds.
Following weight averaging, we refit the temperature parameters for probability calibration.
The AutoThink models use the same Stage 2 data but retain causal attention, preparing them for autoregressive reasoning in Stage 3.

\begin{figure}[!tbp]
\centering
\adjincludegraphics[max width=\linewidth,max height=0.5\textheight]{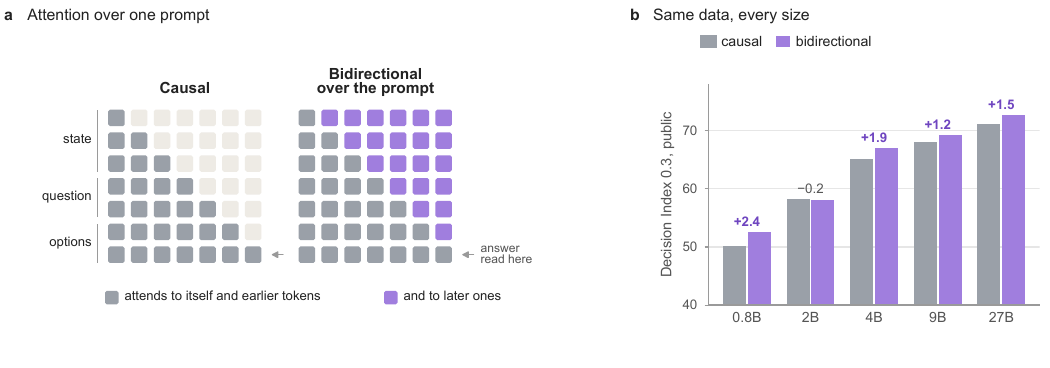}
\vspace{-10mm}
\caption{\textbf{Bidirectional attention improves single-pass decision performance across most evaluated model scales.} (a) Causal and bidirectional attention patterns in full-attention layers, with linear-attention layers remaining causal and decision probabilities read from the final prompt position. (b) DI 0.3 performance of Stage 2 models trained on identical data from the same Stage 1 checkpoint, differing only in their attention configurations. Runs with identical training configurations exhibit variations of 0.2--0.7 index points across random seeds.}
\label{fig:attention-ablation}
\end{figure}

\paragraph{Bidirectional Attention over the Prompt.}
We train the Instant models with bidirectional attention in their full-attention layers, allowing earlier prompt representations to incorporate information from the question and candidate options that follow them.
At 4B and 9B, this modification applies to 8 of the 32 layers, while the linear-attention layers and option-letter readout remain unchanged (Figure~\ref{fig:attention-ablation}a).
We train with the modified attention pattern rather than applying it only at inference, since changing the attention mask of a causally trained model can disrupt its predictions~\citep{behnamghader2024llm2vec}.
Perplexity Decider v1.1 adopts a similar bidirectional attention design.

Controlled comparisons using identical training data and the same Stage 1 checkpoint show that bidirectional attention improves DI 0.3 performance at four of the five evaluated model scales, with gains of up to 2.4 points (Figure~\ref{fig:attention-ablation}b).
Additional evaluations show improvements on DI 0.2.1 and other benchmark-related tasks, providing further evidence of its effectiveness beyond DI 0.3.
These results support bidirectional attention as an effective adaptation for single-pass decision prediction across most evaluated scales and settings.

\subsubsection{Stage 3: Reasoning Supervision and RL}\label{sec:stage3-reasoning-rl}
Stage 3 equips the causal Stage 2 models with explicit reasoning capabilities for decisions requiring multi-step computation, while preserving efficient single-pass prediction.
We first apply reasoning supervision to improve reasoning generation and termination, followed by reinforcement learning to optimize reasoning for decision quality and computational efficiency.
The reasoning process and both probability readouts share model weights.
\paragraph{Reasoning Supervision.}
Decision-oriented training can substantially impair a model's ability to generate and reliably terminate reasoning traces.
In our experiments, the decision-trained 4B model produces little or no reasoning for approximately 40\% of questions on which it is prompted to reason, while the 9B model continues generating until the length limit on approximately 70\% of such questions.
These contrasting failure modes highlight the need to train both reasoning generation and termination, with the latter also posing challenges for general reasoning models~\citep{chen2025not}.
We first train an adapter on Lux-Think, using approximately 4,000--21,000 supervised reasoning examples per model.
Related work has explored supervised training, distillation, and reinforcement learning to develop reasoning capabilities in smaller models~\citep{zhuang2025effective}.
Following reasoning supervision, all evaluated AutoThink models achieve higher post-reasoning accuracy than with single-pass prediction.

Despite these improvements, reasoning supervision does not fully resolve termination failures or excessive reasoning length.
We therefore introduce \rlcdtwo{} to optimize the reasoning policy for probabilistic decision quality, rewarding reasoning-induced improvements in decision probabilities while imposing a soft penalty on overlong reasoning traces.
This final training stage further improves reasoning efficiency while largely preserving decision accuracy, as detailed in Section~\ref{sec:rlcd}.

\section{\rlcdtwo{}: Reinforcement Learning for Calibrated Deliberative Decisions}\label{sec:rlcd}
Reinforcement Learning for Calibrated Decisions (RLCD) was introduced alongside Jev, a hosted decision model whose training procedure has not been publicly disclosed~\citep{almeida2026introducing}.
OpenJev-RLCD subsequently explores a two-stage approach that calibrates post-rationale probability readouts before optimizing reasoning with proper scoring rules~\citep{gao2026openjevrlcd}.
Building on this line of research, we propose \rlcdtwo{}, extending decision-oriented reinforcement learning to explicitly optimize the contribution of reasoning to probabilistic decision quality.
Our formulation combines a frozen probability scorer, a decision-weighted proper scoring rule, and a soft length penalty to balance decision quality with reasoning efficiency.

\subsection{Scoring Foundations and Reward Formulation}\label{sec:proper-scoring-rules}\label{sec:reasoning-based-rlcd}

\paragraph{Proper Scoring Rules.}
We employ proper scoring rules to evaluate the full probability distribution over candidate decisions rather than only the correctness of the selected option.
For a predicted distribution $p=(p_1,\ldots,p_K)$ and a target distribution $t$, a scoring rule $S(p,t)$ is \emph{proper} if its expected value is maximized by reporting the true outcome distribution, and \emph{strictly proper} if this maximizer is unique.

We consider three scoring rules.
The Brier score $S_{\mathrm{B}}(p,t)=-\|p-t\|_2^2$ and logarithmic score $S_{\mathrm{L}}(p,t)=\sum_{k=1}^{K}t_k\log p_k$ evaluate probability quality through squared error and log likelihood, respectively, with the latter corresponding to the negative cross-entropy objective used in Stages 1 and 2.
To account for confidence-sensitive decisions, we further adopt a decision-weighted proper scoring rule that emphasizes probability errors near operational confidence thresholds.
For an option with reported probability $q$ and binary outcome $z\in\{0,1\}$, define
\begin{equation}
\label{eq:decision-weighted-score}
s(q,z)=zW(q)-C(q),\qquad
W(q)=\int_0^q w(c)\,dc,\qquad
C(q)=\int_0^q c\,w(c)\,dc.
\end{equation}
Summing the option-wise scores yields the decision-weighted score $S_{\mathrm{D}}(p,t)=\sum_{k=1}^{K}s(p_k,t_k)$.
For an option with true correctness probability $\pi$, the expected score satisfies
\begin{equation}
\label{eq:decision-weighted-properness}
\frac{\partial}{\partial q}\mathbb{E}[s(q,z)]
=w(q)(\pi-q).
\end{equation}
The expected score is maximized at $q=\pi$ for nonnegative $w$, with a unique maximizer when $w$ is strictly positive over the relevant probability range.
We define $w(c)$ as an equal mixture of a uniform density and a $\mathrm{Beta}(8,2)$ density, placing greater emphasis on high-confidence decisions.
For ordered response levels, the score is applied to cumulative probabilities.

\paragraph{Reasoning Reward.}
Many reasoning-oriented reinforcement learning methods primarily reward answer correctness~\citep{guo2025deepseekr1,ma2025sqlr1,yu2025dapo}.
In contrast, \rlcdtwo{} evaluates changes in probabilistic decision quality following reasoning while accounting for generation cost.
Let $p_0(x)$ denote the initial single-pass probability distribution for input $x$, and $p_1(x,\tau)$ the distribution obtained after reasoning trace $\tau$, both produced by a frozen copy of the starting model.
We define the reasoning reward as
\begin{equation}
\label{eq:rlcd2-reward}
r(x,\tau)=S\big(p_1(x,\tau),t\big)-S\big(p_0(x),t\big)-\lambda\max\big(0,|\tau|-B\big),
\end{equation}
where $S$ is the decision-weighted score used for the released models, $B$ is the soft reasoning-length budget, and $\lambda=10^{-4}$ controls the penalty per token beyond this budget.
The proper-score difference quantifies the change in probabilistic decision quality following reasoning, while the length penalty discourages excessive computation without imposing a hard truncation at $B$.
Related approaches also penalize overlong generations, including DAPO~\citep{yu2025dapo}, while Kimi k1.5 favors shorter responses relative to other samples for the same prompt~\citep{kimiteam2025kimi}.

\subsection{Policy Optimization and Training}\label{sec:rlcd2-optimization}
A frozen copy of the reasoning-supervised starting model provides both probability readouts for reward computation, while all language-model parameters of the reasoning policy are optimized.
Traces terminating within eight tokens are treated as empty reasoning and assigned zero reward.
Training is restricted to questions whose maximum initial option probability is below 0.9, matching the confidence gate used at inference.

We optimize the expected reward using group-relative policy gradients~\citep{shao2024deepseekmath,guo2025deepseekr1}.
For each prompt, we sample $G$ reasoning traces, center their rewards within the group, and clip the resulting advantages as $A_i=\mathrm{clip}(r_i-\bar r,-5,5)$.
The clipped surrogate objective, regularized toward the starting policy, is
\begin{equation}
\label{eq:rlcd2-policy-loss}
\mathcal{L}=-\frac{1}{N}\sum_{i\in\mathcal{T}}\sum_{s=1}^{|\tau_i|}\min\Big(\rho_{i,s}A_i,\;\mathrm{clip}\big(\rho_{i,s},1-\epsilon,1+\epsilon\big)A_i\Big)+\beta\widehat{\mathrm{KL}}\big(\pi_\theta\,\|\,\pi_{\mathrm{ref}}\big).
\end{equation}
Here $\mathcal{T}$ indexes the sampled traces in the training batch, $N$ is the total number of generated tokens, and $\rho_{i,s}$ is the probability ratio between the current and sampling policies for token $s$ in trace $i$.
The KL estimator averages $\pi_{\mathrm{ref}}/\pi_\theta-\log(\pi_{\mathrm{ref}}/\pi_\theta)-1$ over generated tokens.

For groups containing only nonempty traces, the initial decision score is constant across traces and cancels under reward centering.
Consequently, their relative advantages depend on post-reasoning scores and length penalties, while the initial score provides a reference for interpreting the incremental value of reasoning.
Empty traces retain zero reward.
Following~\citet{liu2025understanding}, we do not normalize advantages by the within-group standard deviation, allowing groups with smaller reward differences to contribute proportionally smaller updates.

\subsection{Effects on Reasoning and Decision Quality}\label{sec:rlcd-effects}
We evaluate the incremental effects of \rlcdtwo{} by comparing reasoning-supervised checkpoints with their post-\rlcdtwo{} counterparts on 2,986 held-out examples from the training distribution.
Reasoning is invoked for every example, with one trace per question and a 4,096-token generation limit.
We additionally evaluate 7,800 core DecisionPivot examples under the default confidence gate, using the pair-accuracy metric described in Section~\ref{sec:decisionpivot}.
Table~\ref{tab:reasoning-training-effects} summarizes the held-out results, with default-answer probabilities obtained by combining single-pass and post-reasoning predictions (Section~\ref{sec:model-variants}).

\begin{table}[!tb]
\centering
\caption{\textbf{\rlcdtwo{} improves reasoning termination and efficiency while largely preserving held-out accuracy.} Results compare reasoning-supervised and post-\rlcdtwo{} checkpoints of the 4B, 9B, and 27B AutoThink models on 2,986 held-out examples, after excluding 35 examples overlapping with training or RL data. Reasoning is invoked for every example, with one trace and a 4,096-token generation limit. Accuracy is reported in percent, and default-answer probabilities combine post-reasoning and single-pass predictions with weights of 75\% and 25\%, respectively (Section~\ref{sec:adaptive-reasoning}).}\label{tab:reasoning-training-effects}
{\small\setlength{\tabcolsep}{6pt}\renewcommand{\arraystretch}{1.1}
\noindent\begin{adjustbox}{max width=\linewidth}
\begin{tabular}{lrrrrr}
\toprule
\textbf{Checkpoint} & {\begin{tabular}[b]{@{}r@{}}\textbf{Single-pass}\\\textbf{accuracy}\end{tabular}} & {\begin{tabular}[b]{@{}r@{}}\textbf{Post-reasoning}\\\textbf{accuracy}\end{tabular}} & {\begin{tabular}[b]{@{}r@{}}\textbf{Default-answer}\\\textbf{accuracy}\end{tabular}} & {\begin{tabular}[b]{@{}r@{}}\textbf{Termination}\\\textbf{rate}\end{tabular}} & {\begin{tabular}[b]{@{}r@{}}\textbf{Trace tokens:}\\\textbf{mean /}\\\textbf{90th percentile}\end{tabular}} \\
\midrule
\mdgrouprow \multicolumn{6}{l}{\textbf{4B}} \\
\mdbodyrow After reasoning supervision & 72.04 & 73.91 & 74.35 & 89.5\% & 1,140 / 4,096 \\
\mdourrow \textbf{After \rlcdtwo{}} & 72.10 & 74.61 & 74.82 & 99.3\% & 722 / 1,358 \\
\midrule
\mdgrouprow \multicolumn{6}{l}{\textbf{9B}} \\
\mdbodyrow After reasoning supervision & 79.37 & 79.97 & 80.38 & 45.9\% & 2,494 / 4,096 \\
\mdourrow \textbf{After \rlcdtwo{}} & 79.44 & 80.27 & 80.48 & 93.9\% & 868 / 2,658 \\
\midrule
\mdgrouprow \multicolumn{6}{l}{\textbf{27B}} \\
\mdbodyrow After reasoning supervision & 77.13 & 78.77 & 79.07 & 97.2\% & 313 / 446 \\
\mdourrow \textbf{After \rlcdtwo{}} & 76.99 & 79.17 & 79.54 & 97.4\% & 303 / 435 \\
\bottomrule
\end{tabular}
\end{adjustbox}\par}
\end{table}

\begin{figure}[!tb]
\centering
\vspace{-3mm}
\adjincludegraphics[max width=\linewidth,max height=0.5\textheight]{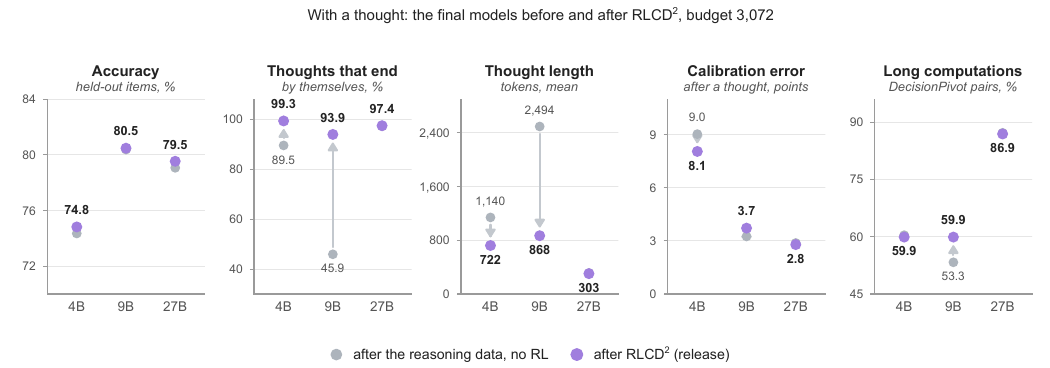}
\vspace{-7mm}
\caption{\textbf{\rlcdtwo{} improves termination and efficiency while largely preserving held-out accuracy.} The 4B, 9B, and 27B AutoThink models before RL (gray) and after \rlcdtwo{} (violet), using the decision-weighted score and a soft penalty budget of 3,072 tokens (distinct from the 4,096-token generation limit). Held-out metrics are measured on 2,986 examples with reasoning invoked for every example: default-answer accuracy, termination rate, mean trace length, and post-reasoning calibration error before temperature scaling. DecisionPivot pair accuracy is measured on 7,800 core items at the default confidence gate of 0.9.}\label{fig:rlcd2-overview}
\end{figure}

\paragraph{Reasoning Termination and Efficiency.}
The primary benefits of \rlcdtwo{} are improved reasoning termination and reduced generation cost (Table~\ref{tab:reasoning-training-effects}; Figure~\ref{fig:rlcd2-overview}).
At 9B, termination increases from 45.9\% to 93.9\%, while mean trace length decreases from 2,494 to 868 tokens, a 65\% reduction.
At 4B, termination rises from 89.5\% to 99.3\%, with mean trace length decreasing by 37\% and the 90th percentile falling from 4,096 to 1,358 tokens.
The 27B model already terminates 97.2\% of traces with a mean length of 313 tokens, leaving limited room for further improvement.

\paragraph{Decision Quality and Calibration.}\label{sec:calibration-resolution-accuracy}
Held-out default-answer accuracy remains largely unchanged across all three scales, with uncertainty intervals for the changes including zero.
At 4B, raw post-reasoning calibration error decreases from 9.0 to 8.1 percentage points, with an interval of $[-1.7,-0.1]$ for the change.
After temperature scaling fitted on development examples, however, the corresponding errors are comparable at 3.1 and 3.2 points.
At 9B and 27B, calibration changes remain within statistical uncertainty, consistent with their lower initial errors.

On the held-out training distribution, the 4B model's gains are primarily reflected in raw calibration rather than answer accuracy or confidence-based ranking, with most of its predictive improvement already established through reasoning supervision.
On DecisionPivot, the 9B model exhibits a different pattern: pair accuracy increases from 53.3\% to 59.9\% under the 3,072-token budget, with an interval of $[+5.0,+8.3]$ for the change (Table~\ref{tab:rlcd2-reward-ablation}).
Together, these results show that \rlcdtwo{} substantially improves reasoning efficiency while largely preserving held-out decision accuracy, with additional gains in calibration or benchmark accuracy depending on the model scale and evaluation setting.
Appendix~\ref{sec:brier-decomposition} further discusses calibration and resolution under proper scoring rules.

\subsection{Reward Design and Reasoning Budget}\label{sec:reasoning-budget-tradeoffs}
\paragraph{Reasoning Budget and Accuracy Trade-offs.}
The soft reasoning budget determines when the length penalty takes effect, governing the trade-off between reasoning efficiency and accuracy on computation-intensive decisions (Table~\ref{tab:rlcd2-reward-ablation}).
Before RL, the 4B AutoThink model generates approximately 2,100 tokens per trace on DecisionPivot and achieves 60.4\% pair accuracy at the default confidence gate.
A 512-token budget approximately halves reasoning length but reduces pair accuracy to 52.9\%.
Increasing the budget to 2,048 and 3,072 tokens restores pair accuracy to 59.2\% and 59.9\%, respectively, with the latter differing from the pre-RL checkpoint by $-0.4$ points (interval $[-1.8,+0.8]$).
Removing the length penalty increases pair accuracy to 61.7\% but produces 40\% longer traces, with approximately one third failing to terminate at the 4,096-token generation limit, compared with about one tenth under the 3,072-token budget.
Across these configurations, held-out accuracy changes by at most 0.6 points, while budgets of 512--3,072 tokens achieve approximately 99\% termination.
At the shorter budget, both Brier and decision-weighted rewards improve calibration and termination but reduce DecisionPivot accuracy, highlighting the cost of overly restrictive reasoning.
\begin{table}[!t]
\centering
\caption{\textbf{Effects of reward design and reasoning budget on decision quality and reasoning efficiency.} Comparison of reasoning-supervised checkpoints and selected reinforcement learning configurations for the 4B, 9B, and 27B AutoThink models. Default-answer accuracy, termination rate, mean trace length, and post-reasoning calibration error are evaluated on 2,986 held-out examples with reasoning invoked for every example; DecisionPivot pair accuracy is measured on 7,800 core items at the default confidence gate of 0.9. Accuracy is reported in percent and calibration error in percentage points, with temperature-scaled values in parentheses. Default-answer probabilities assign 75\% weight to the post-reasoning readout (Section~\ref{sec:adaptive-reasoning}).}\label{tab:rlcd2-reward-ablation}
{\small\setlength{\tabcolsep}{5pt}\renewcommand{\arraystretch}{1.1}
\noindent\begin{adjustbox}{max width=\linewidth}
\begin{tabular}{lrrrrr}
\toprule
\textbf{Checkpoint / RL reward} & {\begin{tabular}[b]{@{}r@{}}\textbf{Default}\\\textbf{accuracy}\end{tabular}} & {\begin{tabular}[b]{@{}r@{}}\textbf{Termination}\\\textbf{rate}\end{tabular}} & {\begin{tabular}[b]{@{}r@{}}\textbf{Mean trace}\\\textbf{tokens}\end{tabular}} & {\begin{tabular}[b]{@{}r@{}}\textbf{Post-reasoning}\\\textbf{calibration error}\end{tabular}} & {\begin{tabular}[b]{@{}r@{}}\textbf{DecisionPivot}\\\textbf{pair accuracy}\end{tabular}} \\
\midrule
\mdgrouprow \multicolumn{6}{l}{\textbf{4B AutoThink model}} \\
\mdbodyrow After reasoning supervision (no RL) & 74.35 & 89.5\% & 1,140 & 9.03 (3.08) & 60.4 \\
\mdbodyrow Brier score, budget 512 & 74.68 & 99.3\% & 504 & 6.96 (3.06) & 53.8 \\
\mdbodyrow Decision-weighted score, budget 512 & 74.58 & 99.7\% & 511 & 6.29 (2.44) & 52.9 \\
\mdbodyrow Decision-weighted score, budget 2,048 & 74.28 & 99.3\% & 677 & 8.11 (2.55) & 59.2 \\
\mdourrow \textbf{Decision-weighted score, budget 3,072 (release)} & 74.82 & 99.3\% & 722 & 8.05 (3.24) & 59.9 \\
\mdbodyrow Decision-weighted score, no length penalty & 74.48 & 91.6\% & 1,186 & 9.38 (2.73) & 61.7 \\
\midrule
\mdgrouprow \multicolumn{6}{l}{\textbf{9B AutoThink model}} \\
\mdbodyrow After reasoning supervision (no RL) & 80.38 & 45.9\% & 2,494 & 3.25 (2.15) & 53.3 \\
\mdbodyrow Decision-weighted score, budget 2,048 & 80.64 & 98.1\% & 601 & 3.29 (2.32) & 61.1 \\
\mdourrow \textbf{Decision-weighted score, budget 3,072 (release)} & 80.48 & 93.9\% & 868 & 3.72 (1.42) & 59.9 \\
\midrule
\mdgrouprow \multicolumn{6}{l}{\textbf{27B AutoThink model}} \\
\mdbodyrow After reasoning supervision (no RL) & 79.07 & 97.2\% & 313 & 2.88 (1.97) & 87.2 \\
\mdourrow \textbf{Decision-weighted score, budget 3,072 (release)} & 79.54 & 97.4\% & 303 & 2.79 (2.00) & 86.9 \\
\bottomrule
\end{tabular}
\end{adjustbox}\par}
\end{table}
\begin{figure}[!htb]
\centering
\vspace{-3mm}
\adjincludegraphics[max width=\linewidth,max height=0.5\textheight]{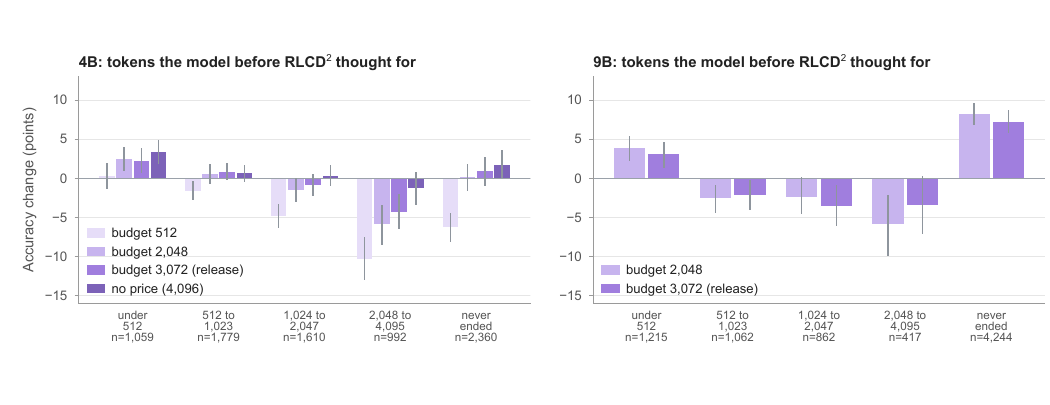}
\vspace{-13mm}
\caption{\textbf{\rlcdtwo{} improves decisions with previously unterminated traces but can reduce accuracy on long computations.} Change in DecisionPivot accuracy at the default gate, after RL minus before RL, grouped by pre-RL trace length. Error bars show 95\% intervals, and $n$ denotes the number of items in each group.}\label{fig:rlcd2-trace-effects}
\end{figure}
\paragraph{Effects on Longer Computations.}
To identify where these trade-offs arise, we group DecisionPivot examples by their pre-RL reasoning lengths (Figure~\ref{fig:rlcd2-trace-effects}).
At 4B, a 2,048-token budget improves accuracy by 2.5 points on examples with pre-RL traces shorter than 512 tokens but reduces accuracy by 5.9 points on those with traces of 2,048--4,095 tokens.
Increasing the budget to 3,072 tokens reduces the loss in the latter group to 4.2 points.

At 9B, \rlcdtwo{} yields gains on previously unterminated traces, which account for 54\% of DecisionPivot examples before RL.
Accuracy on these examples improves by 7.3 points, although examples with pre-RL traces of 2,048--4,095 tokens still experience losses of 3.4 and 5.8 points under budgets of 3,072 and 2,048 tokens, respectively.
The 27B model exhibits reliable termination and short reasoning traces, with no statistically clear changes under \rlcdtwo{}.
We adopt a training budget of 3,072 tokens across the released AutoThink models, which maintains 99.3\% termination for the 4B model on held-out examples while limiting accuracy losses on longer computations.

\paragraph{Termination Dynamics and Reward Components.}
The substantial reduction in reasoning length at 9B is associated with repetitive generation loops that persist until the token limit.
Early in RL training, 29\% of 9B traces reach this limit, compared with 9\% at 4B.
Penalizing overlong traces lowers their rewards and the group mean, increasing the relative advantages of terminating trajectories and encouraging earlier termination.
This pressure diminishes as repetitive loops disappear, particularly at 4B.
In 4B training, rollouts shorter than 16 tokens generally receive negative relative advantages, although their share of the rollouts increases from 1\% to 3\% as termination improves.

Controlled reward comparisons on an earlier checkpoint further distinguish the contributions of the reward components.
Including the length penalty produces nearly universal termination even when the predictive reward is replaced with random noise, whereas removing it results in longer traces and more termination failures.
Calibration improvements, however, depend on the predictive reward: proper-score rewards outperform accuracy-only rewards, while replacing the length penalty with a termination bonus provides no calibration improvement.
These findings indicate that the length penalty primarily drives termination improvements, whereas proper-score rewards contribute to probability calibration.

\begin{table}[t]
\centering
\caption{\textbf{StartLux-Decision achieves leading performance across parameter scales on the public subset of DI 0.3.} Scores on DI 0.3 and DI 0.2.1 are reported in points. Reference scores are drawn from the public leaderboard, except for Decision 3.0, which we evaluate using the same pipeline as StartLux-Decision ($^\ddagger$). The leaderboard excludes models that generate reasoning traces; a dash indicates an unavailable result.}\label{tab:decision-index-results}
{\small\setlength{\tabcolsep}{6pt}\renewcommand{\arraystretch}{1.1}
\noindent\begin{adjustbox}{max width=\linewidth}
\begin{tabular}{lrr}
\toprule
\textbf{System} & {\begin{tabular}[b]{@{}r@{}}\textbf{DI}\\\textbf{0.3}\end{tabular}} & {\begin{tabular}[b]{@{}r@{}}\textbf{DI}\\\textbf{0.2.1}\end{tabular}} \\
\midrule
\mdgrouprow \multicolumn{3}{l}{\textbf{27B}} \\
\mdourrow StartLux-Decision-27B-AutoThink & \textbf{73.37} & \textbf{72.60} \\
\mdourrow StartLux-Decision-27B-Instant & 72.83 & 72.12 \\
\mdbodyrow Decision 3.0 d3 (vLLM Semantic Router)$^\ddagger$ & 65.44 & 64.77 \\
\mdbodyrow Perplexity Decider v1.1 & 62.25 & -- \\
\mdbodyrow Jev (hosted) & 57.96 & 57.91 \\
\midrule
\mdgrouprow \multicolumn{3}{l}{\textbf{9B}} \\
\mdourrow StartLux-Decision-9B-AutoThink & \textbf{70.45} & \textbf{69.65} \\
\mdourrow StartLux-Decision-9B-Instant & 69.96 & 69.04 \\
\mdbodyrow Decision 3.0 d3-flash (vLLM Semantic Router)$^\ddagger$ & 57.88 & 57.08 \\
\mdbodyrow Bespoke Nimble 9B v3 & 57.19 & -- \\
\midrule
\mdgrouprow \multicolumn{3}{l}{\textbf{4B}} \\
\mdourrow StartLux-Decision-4B-AutoThink & \textbf{68.21} & \textbf{67.42} \\
\mdourrow StartLux-Decision-4B-Instant & 67.88 & 66.73 \\
\mdbodyrow Decision 3.0 d3-mini (vLLM Semantic Router)$^\ddagger$ & 54.93 & 54.11 \\
\mdbodyrow d1-3B (Liquid AI) & 48.99 & -- \\
\midrule
\mdgrouprow \multicolumn{3}{l}{\textbf{2B}} \\
\mdourrow StartLux-Decision-2B-Instant & \textbf{59.14} & \textbf{57.80} \\
\mdbodyrow Decision 3.0 d3-nano (vLLM Semantic Router)$^\ddagger$ & 42.24 & 41.23 \\
\mdbodyrow Decision 2.0 Sol (vLLM Semantic Router) & 30.05 & -- \\
\midrule
\mdgrouprow \multicolumn{3}{l}{\textbf{0.8B}} \\
\mdourrow StartLux-Decision-0.8B-Instant & \textbf{53.75} & \textbf{52.37} \\
\mdbodyrow Decision 3.0 d3-lite (vLLM Semantic Router)$^\ddagger$ & 36.89 & 35.97 \\
\mdbodyrow Decision 3.0 d3-edge, 0.6B (vLLM Semantic Router)$^\ddagger$ & 28.77 & 27.95 \\
\mdbodyrow OneJev 0.8B (OmniJev) & 22.22 & -- \\
\bottomrule
\end{tabular}
\end{adjustbox}\par}
\end{table}

\section{Experimental Evaluation}\label{sec:experimental-evaluation}

We evaluate decision accuracy and probability reliability across tasks with varying computational demands.
Decision Index 0.2.1 and 0.3 assess performance on established benchmarks, while DecisionPivot examines computation-intensive decisions and sensitivity to answer-changing details through programmatically verifiable contrast pairs.
We further evaluate probability calibration.

\subsection{Experimental Setup}\label{sec:experimental-setup}
All StartLux-Decision models are evaluated under fixed inference configurations, including the prompt format, probability readout, and checkpoint-specific calibration temperatures described in Section~\ref{sec:model-design}.
Neither model parameters nor calibration temperatures are fitted on the evaluation sets (Appendix~\ref{sec:D}).
Decision Index (DI) 0.2.1 aggregates chance-corrected scores across 38 benchmarks spanning five domains.
DI 0.3 removes ForecastBench and replaces the original GSM8K questions with 2,638 reconstructed requests designed to eliminate answer-revealing cues in the candidate options.
The public evaluation accounts for 20\% of the DI 0.3 leaderboard score, while private evaluations covering existing capabilities and additional domains determine the remaining 80\%.
We evaluate only the public subset, with most evaluated tasks belonging to families represented during training.
The public subset of DI also served as the development benchmark.
We compared Stage 2 data recipes, prompt attention variants and the seed averages that form the Instant checkpoints by their DI public scores, so these scores reflect model selection and are not independent test results.
Calibration temperatures and the AutoThink gate, mixture weight and reasoning budget were set on held-out items of the training distribution, not on DI.
DecisionPivot was not used for any data, checkpoint or configuration choice, and it provides the independent evaluation in this report.

\subsection{Decision Index}\label{sec:decision-index-results}
\begin{figure}[!t]
\centering
\adjincludegraphics[max width=\linewidth,max height=0.5\textheight]{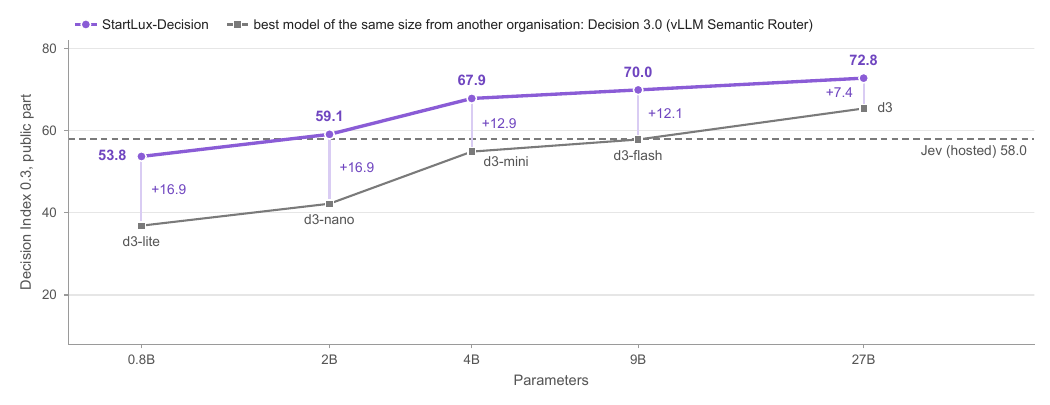}
\caption{\textbf{The Instant models outperform comparable-scale reference decision models by 7.4--16.9 points on the public subset of DI 0.3.} Scores are plotted against parameter count. The gray line connects the strongest evaluated third-party model at each scale, all from the Decision 3.0 family of vLLM Semantic Router. Jev is hosted, with an undisclosed parameter count.}\label{fig:decision-index-scaling}
\end{figure}

Table~\ref{tab:decision-index-results} compares StartLux-Decision with representative reference models on both editions of Decision Index.
Across all five parameter scales, the Instant models outperform the strongest evaluated comparable-scale third-party decision models, all from the Decision 3.0 family, by 7.4--16.9 points on the public subset of DI 0.3 (Figure~\ref{fig:decision-index-scaling}).
Notably, the 4B Instant model outperforms all evaluated reference decision models, including substantially larger systems.
The relative ordering of StartLux-Decision models remains consistent across DI 0.3 and DI 0.2.1.

\begin{figure}[!t]
\centering
\adjincludegraphics[max width=\linewidth,max height=0.5\textheight]{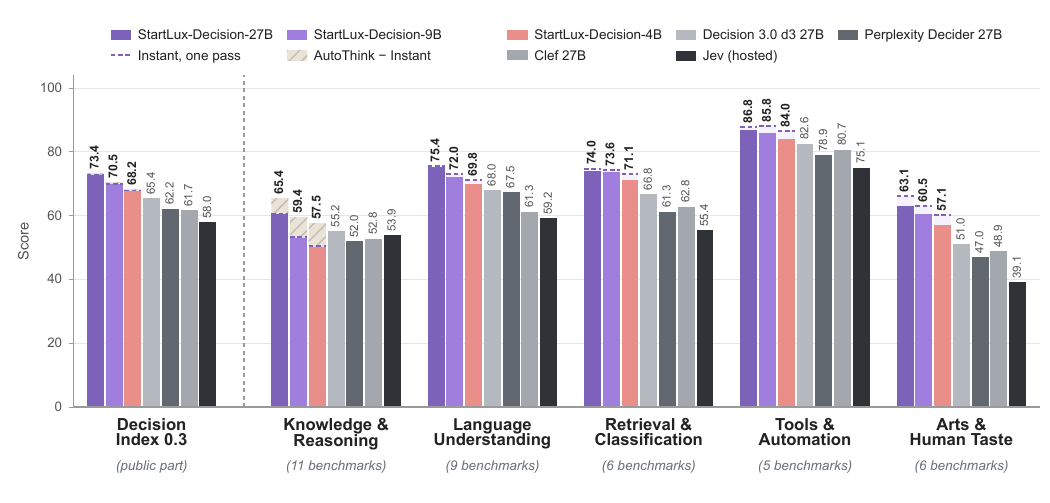}
\caption{\textbf{The 27B Instant model outperforms all evaluated reference decision models across the five DI 0.3 domains.} Overall and domain scores are reported in points on the public subset, with domain weights matching the leaderboard. Solid bars show Instant performance, while hatched extensions represent the score differences between AutoThink and Instant at each scale. When AutoThink scores lower, the dashed line marks the higher Instant score. Reference scores come from the public leaderboard, except for Decision 3.0 d3, which we evaluate using the same pipeline as StartLux-Decision.}\label{fig:decision-index-domains}
\end{figure}

The 27B Instant model outperforms all evaluated reference decision models across the five domains, while the 4B and 9B Instant models lead in four.
Knowledge \& Reasoning is particularly challenging for single-pass prediction, as many questions require explicit computation.
In this domain, the 4B and 9B AutoThink models outperform their Instant counterparts by 7.0 and 6.2 points, respectively (Figure~\ref{fig:decision-index-domains}).
Detailed results for individual Decision Index benchmarks are provided in Table~\ref{tab:D.1} of Appendix~\ref{sec:D}.

\begin{figure}[!t]
\centering
\adjincludegraphics[max width=\linewidth,max height=0.5\textheight]{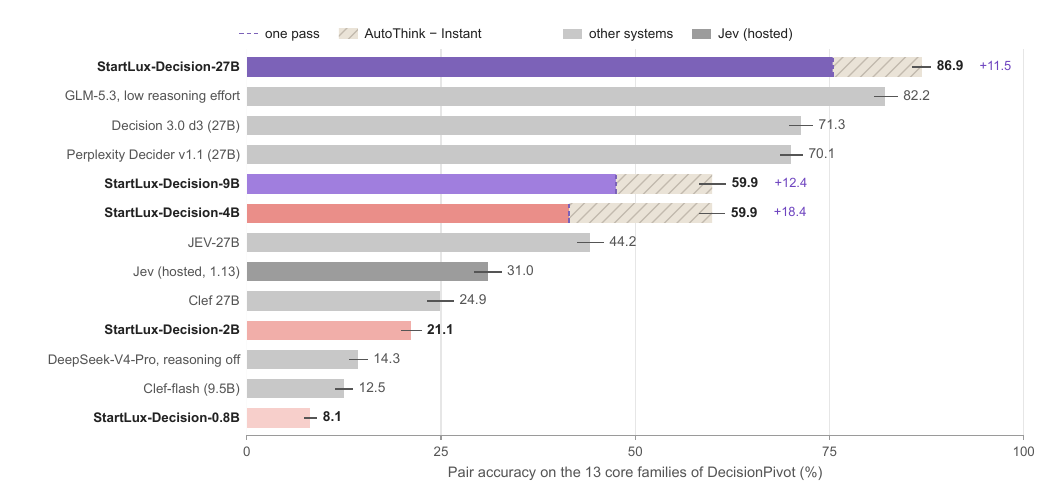}
\caption{\textbf{The 27B AutoThink model achieves the highest pair accuracy among evaluated systems on DecisionPivot.} Results cover 7,800 examples across 13 programmatically generated task families, with 95\% bootstrap confidence intervals. Solid bars show single-pass performance, while hatched extensions indicate the gains of AutoThink over Instant at the default confidence threshold of 0.9.}\label{fig:decisionpivot-overall}

\end{figure}

\begin{figure}[!t]
\centering
\adjincludegraphics[max width=0.99\linewidth,max height=0.5\textheight]{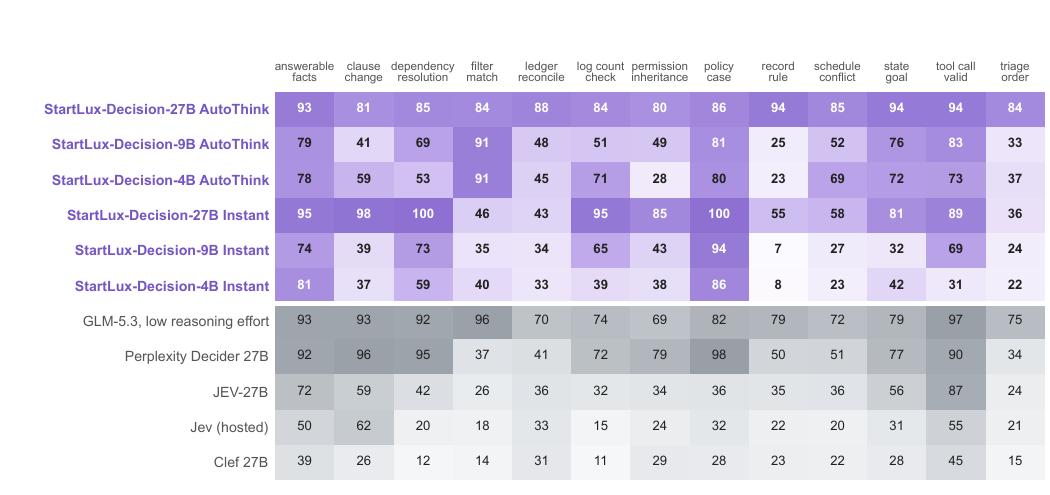}
\caption{\textbf{AutoThink excels on computation-intensive decisions, while strong single-pass models remain competitive on structural tasks.} Pair accuracy is reported in percent across the 13 DecisionPivot core families, with darker cells indicating higher scores. StartLux-Decision models are shown in violet and reference systems in gray.}\label{fig:decisionpivot-families}
\end{figure}

\subsection{DecisionPivot}\label{sec:decisionpivot}
DecisionPivot~\citep{startlux2026decisionpivot} complements Decision Index by evaluating demanding decisions through programmatically verifiable contrastive examples.
Its core track comprises 13 generated task families with 600 examples each, organized predominantly into contrast pairs that differ in a single answer-changing detail.
Pair accuracy requires both examples to be answered correctly and is averaged across the 13 families.
A separate \emph{reused tasks} reference track comprises 35 public test sets, restricted to 34,104 examples absent from our training data.

The 27B Instant model achieves 75.4\% pair accuracy, outperforming Decision 3.0 d3 (71.3\%), Perplexity Decider v1.1 (70.1\%), and the hosted Jev (31.0\%), while GLM-5.3 reaches 82.2\% with explicit reasoning (Figure~\ref{fig:decisionpivot-overall}).
AutoThink substantially improves pair accuracy relative to Instant across all three evaluated scales, with gains of 18.4, 12.4, and 11.5 points at 4B, 9B, and 27B, respectively.
At the default confidence threshold of 0.9, both the 4B and 9B AutoThink models achieve 59.9\% pair accuracy, while the 27B model reaches 86.9\%, outperforming all evaluated reference systems, including GLM-5.3.
These gains are particularly pronounced on computation-intensive tasks.

The family-level results reveal distinct strengths across computational and structural decision tasks (Figure~\ref{fig:decisionpivot-families}).
Relative to their Instant counterparts, the 4B and 9B AutoThink models improve pair accuracy by approximately 51 and 56 points on filtering tasks, 30 and 44 on goal-state verification, and 46 and 25 on schedule-conflict detection, respectively.
In contrast, single-pass prediction remains highly effective for tasks involving policy interpretation, clause changes, and dependency resolution.

\subsection{Probability Calibration}\label{sec:probability-calibration}
\begin{figure}[!t]
\centering
\adjincludegraphics[max width=\linewidth,max height=0.5\textheight]{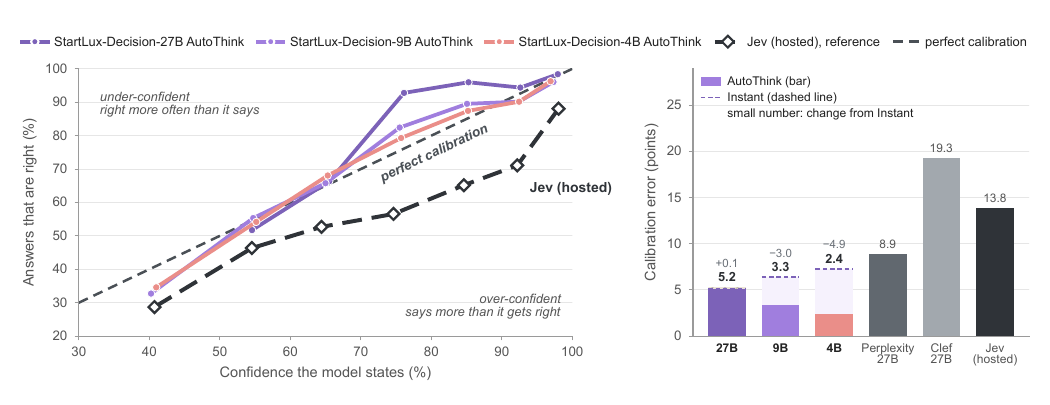}

\caption{\textbf{StartLux-Decision achieves lower calibration error than evaluated reference decision models on DecisionPivot.} Left: empirical accuracy versus predicted confidence (maximum option probability) for the 4B, 9B, and 27B AutoThink models and hosted Jev across 7,800 core examples. Points below the diagonal indicate overconfidence. Right: calibration error based on confidence--accuracy gaps across 15 bins, reported in percentage points. Bars show AutoThink at the default confidence threshold of 0.9, with dashed lines marking the corresponding Instant models.}\label{fig:probability-calibration}

\end{figure}

Reliable decision-making requires well-calibrated probabilities, particularly when downstream systems use confidence to act or defer.
On 7,800 DecisionPivot core examples, we compare predicted confidence, defined as the maximum option probability, with empirical accuracy across 15 confidence bins (Figure~\ref{fig:probability-calibration}).

The evaluated StartLux-Decision models exhibit lower calibration error than the reference systems shown in Figure~\ref{fig:probability-calibration}.
The AutoThink models achieve calibration errors of 2.4, 3.3, and 5.2 percentage points at 4B, 9B, and 27B, respectively, compared with 7.3, 6.4, and 5.1 points for their Instant counterparts.
The 27B AutoThink model is the only StartLux-Decision model that is underconfident on average.

Perplexity Decider v1.1, hosted Jev, and Clef exhibit calibration errors of 8.9, 13.8, and 19.3 points, respectively, with systematic overconfidence.
Previous studies have examined Jev's calibration on datasets~\citep{deusser2026evaluating} and its probabilities for individual decisions~\citep{porcedda2026jev}.
DecisionPivot extends this evaluation to programmatically verifiable decisions involving explicit computation.
These results highlight probability reliability alongside predictive accuracy, while Section~5 further examines its behavior under selection and routing.

\section{Capability and Reliability Analysis}\label{sec:decision-reasoning-analysis}
Beyond overall benchmark performance, we investigate how training choices and adaptive reasoning shape decision performance, and assess the reliability of the resulting probability estimates.
We further examine how downstream selection and routing affect confidence reliability and how policy-aware calibration addresses these effects.

\subsection{Effects of Training Data and Attention Design}\label{sec:training-attention-effects}
To distinguish gains from task-family exposure and attention adaptation, we evaluate additional tasks beyond Decision Index, whose public part also guided development choices (Section~\ref{sec:experimental-setup}).
Additional controlled training runs use the 4B model; analyses of the released AutoThink models cover the 4B, 9B, and 27B scales.

\paragraph{Effects of Training Data.}
Additional decision-training data primarily improve performance on represented task families.
On the reused tasks of DecisionPivot, adding Stage 2 data from rating, safety, preference, and logic families improves the 18 corresponding tasks by 8.6 points, compared with only 0.3 points on tasks outside those families.
Across the evaluated data groups, we find no clear improvement outside their respective task-family coverage.
A similar pattern emerges for procedural decisions: adding related training families raises the 4B model's accuracy on generated answerable-fact decisions from 61.5\% to 75.8\%.
These results indicate task-family specialization, with limited evidence of transfer to unrepresented families.

\paragraph{Effects of Bidirectional Attention.}
Bidirectional prompt attention improves performance beyond task families represented in Stage 2 training.
With a narrower Stage 2 mixture, it improves performance by 3.0 points on 16 tasks whose families are absent from that mixture.
This suggests that attention adaptation can improve single-pass prediction beyond gains attributable to task-family coverage.

\begin{figure}[!t]
\centering
\adjincludegraphics[max width=\linewidth,max height=0.5\textheight]{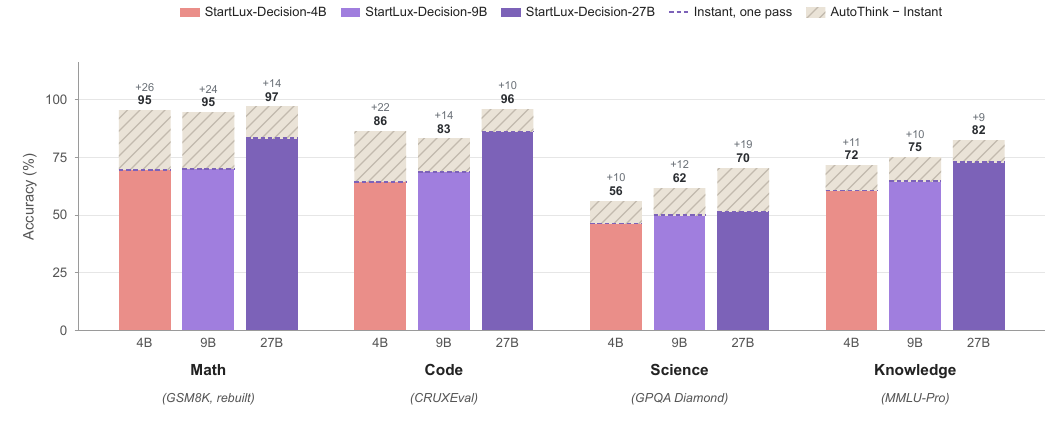}
\caption{\textbf{AutoThink outperforms Instant on computation-intensive tasks across all three evaluated scales.} Solid bars show Instant accuracy, while hatched extensions indicate accuracy differences between the complete model variants. AutoThink uses the default confidence threshold of 0.9. Results cover the 4B, 9B, and 27B models on 2,638 reconstructed GSM8K problems from DI 0.3, along with CRUXEval, GPQA Diamond, and MMLU-Pro from DI 0.2.1 (Table~\ref{tab:D.1}).}\label{fig:computation-task-results}

\end{figure}

\subsection{Adaptive Reasoning and Prediction Fusion}\label{sec:adaptive-reasoning}
\paragraph{Benefits of Adaptive Reasoning.}
AutoThink exhibits its largest advantages over Instant on tasks requiring multi-step computation (Figure~\ref{fig:computation-task-results}).
At 4B, the AutoThink model exceeds its Instant counterpart by 26 accuracy points on reconstructed GSM8K problems, 22 on CRUXEval, 10 on GPQA Diamond, and 11 on MMLU-Pro.
The pattern persists at larger scales.
At 9B, AutoThink exceeds Instant by 24.4 points on GSM8K and 14.4 on CRUXEval, with a 0.5-point overall DI 0.3 advantage.
At 27B, Instant already achieves 83\% accuracy on reconstructed GSM8K problems, with AutoThink scoring another 13.6 points higher.
AutoThink also leads by 19 points on GPQA Diamond, 10 on CRUXEval, and 9 on MMLU-Pro, while increasing DI 0.3 by 0.5 points.
These comparisons involve complete model variants; the following analysis isolates the effect of reasoning within fixed AutoThink checkpoints.

\begin{figure}[!t]
\centering
\adjincludegraphics[max width=\linewidth,max height=0.5\textheight]{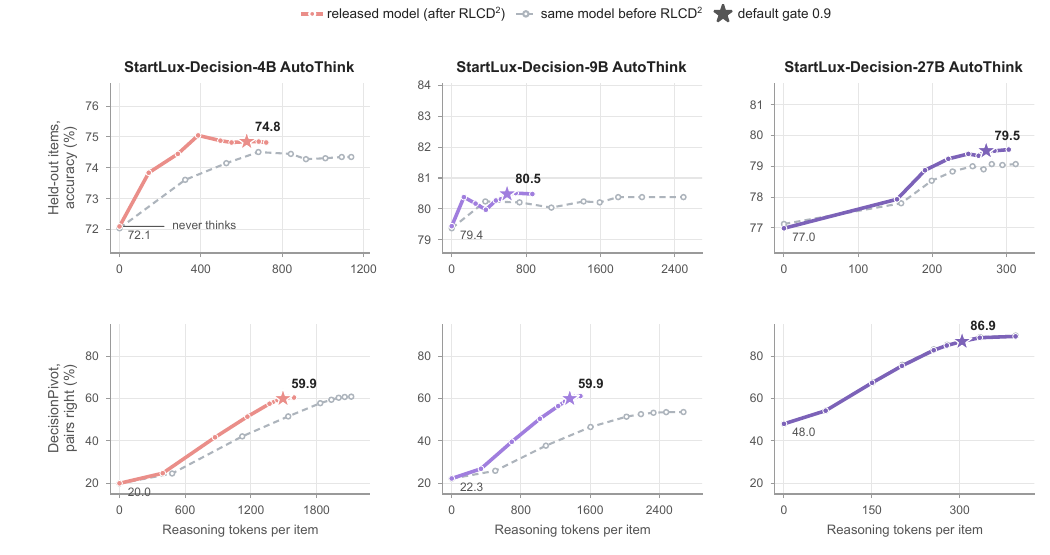}
\caption{\textbf{Confidence-guided reasoning captures most of the attainable accuracy gains, while \rlcdtwo{} reduces reasoning cost at 4B and 9B.} Accuracy is plotted against average reasoning tokens per example, including those answered without reasoning, as the confidence threshold varies from 0 (never reason) to 1 (always reason). Stars indicate the default threshold of 0.9. The top row evaluates 2,986 held-out examples from the training distribution; the bottom row reports pair accuracy on 7,800 DecisionPivot core examples. Each model scale uses its own token axis. Solid curves represent the released AutoThink models after \rlcdtwo{}, while dashed gray curves represent their pre-\rlcdtwo{} checkpoints.}\label{fig:adaptive-reasoning-tradeoffs}
\end{figure}

\paragraph{Confidence-Guided Reasoning.}
Within each AutoThink checkpoint, varying the confidence threshold from 0 (never reason) to 1 (always reason) reveals the accuracy--computation trade-off (Figure~\ref{fig:adaptive-reasoning-tradeoffs}).
Routing low-confidence predictions to reasoning captures most of the attainable accuracy gains.
On held-out examples from the training distribution, the 9B model gains 1.0 accuracy point over its single-pass readout at the default threshold of 0.9, averaging 596 reasoning tokens per example; always invoking reasoning yields no further gain.
At 27B, the same threshold yields a 2.5-point gain with 273 reasoning tokens per example.

The gains are more pronounced on DecisionPivot, where decisions often require explicit computation.
At the default threshold, the 9B model achieves 59.9\% pair accuracy with 1,365 reasoning tokens per example, compared with 22.3\% when reasoning is disabled in the same checkpoint.
The corresponding 27B results are 86.9\% pair accuracy with 304 tokens per example, versus 48.0\% without reasoning.
\rlcdtwo{} further improves the accuracy--efficiency trade-off at smaller scales.
At 9B, it raises DecisionPivot pair accuracy from 53.3\% to 59.9\% while reducing reasoning-token consumption by 40\%; at 4B, token consumption decreases by 25\% with pair accuracy changing by less than 0.5 points.
The 27B model shows little change, consistent with its short reasoning traces (Section~\ref{sec:reasoning-budget-tradeoffs}).

\paragraph{Combining Single-Pass and Post-Reasoning Predictions.}
For questions routed to reasoning, AutoThink combines initial single-pass and post-reasoning probability distributions through a weighted average.
The fusion weight is particularly important for computation-intensive decisions, where initial predictions can be confidently incorrect.
On DecisionPivot, the 9B model achieves 56.6\% pair accuracy with equal weights, 59.9\% with a post-reasoning weight of 75\%, and 61.5\% using only post-reasoning probabilities.

However, maximizing benchmark accuracy does not necessarily optimize probability quality.
On held-out examples from the training distribution, initial predictions are already accurate, whereas post-reasoning predictions exhibit some overconfidence, consistent with prior observations~\citep{lacombe2025dont}.
We therefore select the fusion weight by minimizing the default prediction's Brier score on held-out data.
A post-reasoning weight of 75\% minimizes this score at 9B and 27B and is close to optimal at 4B, motivating the default 75\%--25\% combination.

Relative to equal weighting, the selected fusion improves DecisionPivot pair accuracy by 2.7 points at 4B but reduces held-out accuracy by 0.5 points at 9B and 0.3 points at 4B.
DI 0.3 scores change by less than 0.15 points.
At 27B, increasing the post-reasoning weight to 75\% reduces held-out calibration error from 3.7 to 1.6 points, with little change in accuracy.
In contrast, on Decision Index, the same weight increases calibration error by 0.8--2.0 points relative to equal weighting, according to the leaderboard's calibration metric.
These results highlight the importance of selecting fusion weights based on both probability quality and accuracy under the intended evaluation distribution.

\paragraph{Computational Cost and Remaining Limitations.}
Adaptive reasoning incurs additional generation cost for routed decisions.
At the default threshold, the 4B, 9B, and 27B AutoThink models invoke reasoning on approximately 33\%, 29\%, and 28\% of Decision Index questions, averaging 372, 271, and 110 reasoning tokens per question, respectively.
These averages include questions answered directly without reasoning.
Instant remains preferable for latency-sensitive applications, as it avoids autoregressive generation (Section~\ref{sec:model-variants}).

\subsection{Evidence Sensitivity and Probability Consistency}\label{sec:evidence-probability-consistency}
We examine whether decision probabilities respond to relevant evidence and remain consistent across related question formulations.
Both analyses use single-pass readouts, with reasoning disabled for AutoThink.

\begin{figure}[!t]
\centering
\vspace{-8mm}
\adjincludegraphics[max width=\linewidth,max height=0.5\textheight]{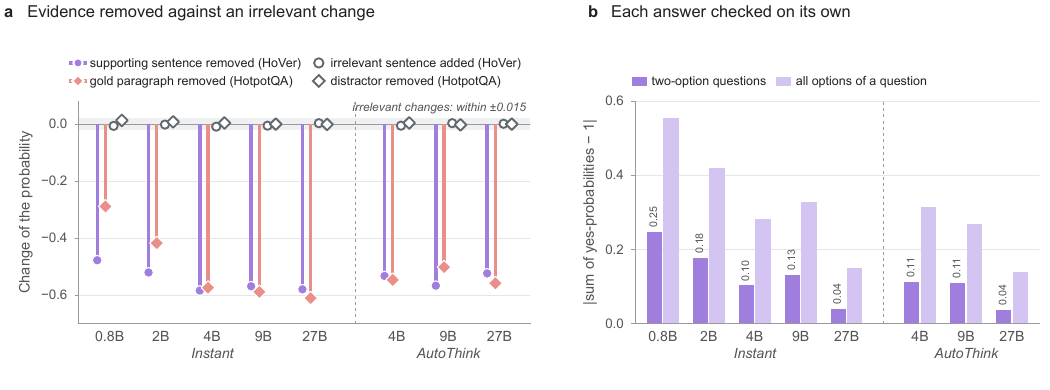}
\vspace{-3mm}
\caption{\textbf{Decision probabilities respond to supporting evidence and become more consistent across related question formulations as model scale increases.} Results use single-pass readouts of the released models. (a) Mean change in the supported answer's probability after removing a supporting sentence from HoVer or a gold paragraph from HotpotQA. Control perturbations add an irrelevant sentence or remove a distractor paragraph, producing mean changes within 0.015 of zero. (b) Mean absolute deviation of the summed affirmative probabilities from one when each candidate option is independently reformulated as a yes-or-no question, reported for binary questions and for the full set of questions.}\label{fig:evidence-consistency}
\end{figure}

\paragraph{Evidence Sensitivity.}
We assess how predicted probabilities respond to removing supporting evidence.
For HoVer claims, we remove one supporting sentence; for HotpotQA questions reformulated as binary decisions, we remove one gold supporting paragraph.
Both probes use evaluation items only, and no evaluation data is used in training.
Across model scales, removing supporting evidence reduces the probability assigned to the supported answer in 90--97\% of cases, with mean decreases ranging from 0.29 to 0.61.
By contrast, adding an irrelevant sentence or removing a distractor paragraph changes the corresponding probability by less than 0.015 on average (Figure~\ref{fig:evidence-consistency}a).
These results indicate sensitivity to supporting evidence with limited responses to irrelevant perturbations.

\paragraph{Probability Consistency.}
We assess whether predicted probabilities remain coherent across related question formulations.
For questions with mutually exclusive and collectively exhaustive options, the affirmative probabilities obtained by asking about each option separately as a yes-or-no decision should ideally sum to one.
Following the consistency test applied to Jev by \citet{li2026beyond}, we measure the mean absolute deviation of this sum from one for binary questions and extend the analysis to questions with more than two options.
Consistency generally improves with model scale: the mean deviation decreases from 0.25 for the 0.8B Instant model to 0.04 for the 27B Instant model.
AutoThink exhibits comparable deviations to Instant at 4B (0.11 versus 0.10) and 27B (0.04 for both), with a lower deviation at 9B (0.11 versus 0.13).

\subsection{Policy-Induced Miscalibration}\label{sec:selection-induced-miscalibration}
Well-calibrated individual predictions need not remain calibrated after downstream selection or routing, as these policies condition returned predictions on confidence and other decision signals.
We examine best-of-$N$, think-gate, cascade, and select-then-think policies.
Using StartLux-Decision-9B as a verifier reveals a substantial selection-induced calibration gap.
Among individual responses assigned 80--90\% confidence, 92\% are correct.
After selecting the highest-scoring response from 32 candidates, accuracy within the same confidence range falls to 78\%.
Calibrating individual responses does not eliminate this effect: selected-response accuracy at fixed confidence declines as the candidate pool grows (Figure~\ref{fig:policy-aware-calibration}a).

Selection changes the conditional distribution of returned predictions.
Choosing the highest-scoring candidate can favor upward fluctuations in confidence estimates, an effect related to the \emph{optimizer's curse}.
Competing candidates' scores also convey information about problem difficulty that the winning score alone may not capture.
Consequently, identical reported confidence can correspond to different empirical accuracies under different selection policies.
The same concern extends to confidence-dependent routing, including AutoThink's think-gate and cascades that escalate uncertain decisions to larger models~\citep{gupta2024language}.
Calibration and acceptance thresholds must therefore be evaluated on the outputs of the deployed policy.
Appendix~\ref{sec:C.2} formalizes these effects.

\paragraph{Selection and Routing Effects.}\label{sec:selection-routing-effects}
We evaluate selection-induced miscalibration across generator--verifier combinations.
After conventional calibration on individual responses, selected responses are overconfident in 111 of 113 generator--verifier pairs at $N=16$.
The calibration gap increases with candidate count in all 126 evaluated runs.
Confidence-based routing exhibits a related effect: for 11 of 12 models, think-gate routing is associated with lower accuracy at matched reported confidence.
The direction of this shift depends on the task, as low initial confidence can identify decisions that benefit substantially from additional computation on reasoning-intensive tasks.

Selection also affects confidence-based acceptance thresholds.
We certify thresholds using a fixed-sequence Clopper--Pearson test at 95\% on the calibration half of each of 20 source-problem splits (Appendix~\ref{sec:C.3}).
Thresholds certified on individual predictions need not preserve their error bounds after selection.
For best-of-16 with a target error rate of 5\%, individually certified thresholds yield observed error rates above the target in 44 of 67 generator--verifier pairs.
By contrast, no such violations are observed when thresholds are certified on policy-selected responses.
These findings support policy-specific threshold certification.

\begin{figure}[!t]
\centering
\vspace{-8mm}
\adjincludegraphics[max width=\linewidth,max height=0.5\textheight]{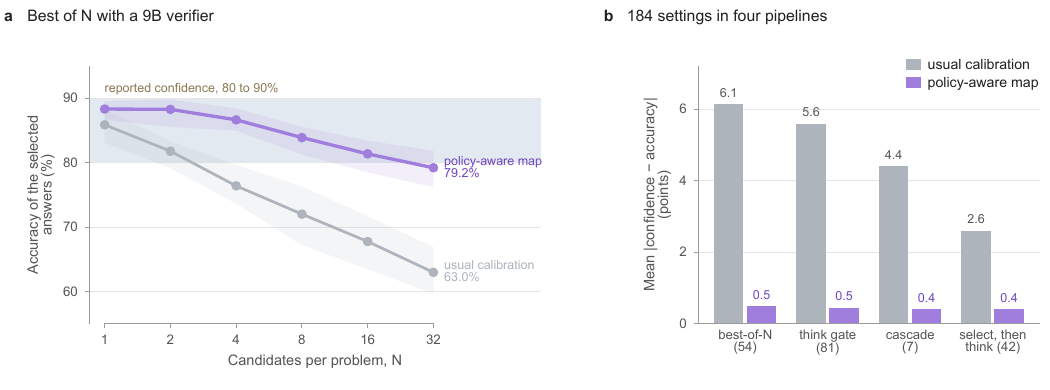}
\caption{\textbf{Selection and routing induce miscalibration that policy-aware calibration mitigates without retraining the underlying models.} (a) Accuracy of responses selected by a 9B verifier from $N$ candidates, restricted to reported confidence of 80--90\%. Shaded bands indicate variation across 20 splits. (b) Mean absolute confidence--accuracy gaps across 184 settings spanning best-of-$N$, think-gate, cascade, and select-then-think pipelines. Policy-aware calibration reduces these gaps across all four pipelines, with signed gaps ranging from $-2.7$ to $+2.1$ percentage points.}\label{fig:policy-aware-calibration}
\end{figure}

\subsection{Policy-Aware Calibration}\label{sec:policy-aware-calibration}\label{sec:policy-calibration-evaluation}
To mitigate policy-induced miscalibration, we calibrate confidence conditional on the information used for selection and routing.
We apply the deployed policy to labeled calibration examples and fit calibration maps using the returned predictions, their reported confidence, and the associated selection or routing scores.
For confidence-based gates and cascades, predictions are stratified by routing-confidence quantiles, with isotonic calibration fitted within each stratum.
For candidate selection, we fit a logistic model using reported confidence, mean candidate score, candidate count $N$, and $\log N$, followed by isotonic calibration of its predictions.
This procedure calibrates reported confidence without modifying the underlying models, while acceptance thresholds are certified separately for the relevant policy routes (Appendices~\ref{sec:C.4} and~\ref{sec:C.5}).

Across 184 settings spanning best-of-$N$, think-gate, cascade, and select-then-think pipelines, policy-aware calibration reduces signed confidence--accuracy gaps from $[-11.7,+20.5]$ to $[-2.7,+2.1]$ percentage points (Figure~\ref{fig:policy-aware-calibration}b).
Mean absolute gaps decrease across all four pipelines.
Nevertheless, calibration remains sensitive to distribution shift: maps fitted on the held-out distribution exhibit residual overconfidence of up to 6.1 percentage points when transferred to unseen datasets.

We further examine calibration and selective prediction under the default AutoThink inference policy.
On 2,986 held-out examples, default-answer calibration errors are 3.7, 2.2, and 1.6 points at 4B, 9B, and 27B, respectively, compared with post-reasoning errors of 8.1, 3.7, and 2.8 points.
At 4B, post-reasoning predictions exhibit 4.0 points of overconfidence despite well-calibrated single-pass predictions, motivating route-specific threshold certification.
At 9B and 27B, both inference routes exhibit signed confidence--accuracy gaps within 1.5 points.
Under a target error rate of 5\%, approximately half of the default predictions from the 9B and 27B models can be accepted, compared with 2\% at 4B.
For the 4B model, restricting acceptance to post-reasoning predictions increases coverage to 23\%, reflecting errors among highly confident single-pass predictions.
These findings distinguish probability calibration from selective coverage, which additionally depends on discrimination.

\begin{table}[!t]
\centering
\vspace{-3mm}
\caption{\textbf{Single-pass decision readouts on an H200 take 31 ms at 0.8B and under 100 ms at 27B for 1,024-token prompts.} Median latency in milliseconds for a single forward pass reading option-letter logits, measured with vLLM 0.31 on one H200. Timings include tokenization and are based on ten measurements after three warm-up passes, with one request in flight. Approximately 90\% of DI requests contain fewer than 1,000 tokens. AutoThink columns report causal readouts without reasoning generation under the same conditions as Instant. BF16 baseline rows are shaded.}\label{tab:vllm-readout-latency}
{\small\setlength{\tabcolsep}{8pt}\renewcommand{\arraystretch}{1.1}
\noindent\begin{tabular}{lrrrrrrrrr}
\toprule
\textbf{Precision} & \textbf{Prompt length} & \multicolumn{5}{c}{\textbf{Instant}} & \multicolumn{3}{c}{\textbf{AutoThink}} \\
\cmidrule(lr){3-7}\cmidrule(lr){8-10}
 & \textbf{(tokens)} & \textbf{0.8B} & \textbf{2B} & \textbf{4B} & \textbf{9B} & \textbf{27B} & \textbf{4B} & \textbf{9B} & \textbf{27B} \\
\midrule
\mdourrow \textbf{BF16} & 256 & 24 & 24 & 31 & 33 & 52 & 29 & 30 & 49 \\
\mdourrow  & 1,024 & 31 & 32 & 41 & 42 & 98 & 39 & 39 & 95 \\
\mdourrow  & 4,096 & 35 & 38 & 72 & 114 & 358 & 70 & 109 & 346 \\
\mdourrow  & 8,192 & 47 & 69 & 142 & 230 & 737 & 131 & 214 & 698 \\
\midrule
\mdbodyrow FP8 & 256 & 26 & 25 & 30 & 29 & 51 & 30 & 30 & 49 \\
\mdbodyrow  & 1,024 & 33 & 33 & 39 & 40 & 72 & 38 & 38 & 70 \\
\mdbodyrow  & 4,096 & 37 & 36 & 61 & 85 & 251 & 56 & 79 & 239 \\
\mdbodyrow  & 8,192 & 43 & 59 & 119 & 172 & 534 & 105 & 154 & 489 \\
\bottomrule
\end{tabular}\par}
\end{table}

\section{Efficient Inference and Deployment}\label{sec:inference-deployment}
We evaluate runtime compatibility, serving latency, local inference, and reduced-precision deployment for StartLux-Decision.
Tables~\ref{tab:vllm-readout-latency} and~\ref{tab:mlx-readout-latency} report single-readout latency without reasoning generation.
Tables~\ref{tab:vllm-model-comparison} and~\ref{tab:llamacpp-model-comparison} compare per-request latency on matched single-pass requests, using one GPU and one request in flight.
For AutoThink, end-to-end latency includes reasoning generation and the post-reasoning readout when invoked.

\subsection{Runtime Support}\label{sec:inference-engines}
Both Instant and AutoThink support inference on vLLM, llama.cpp with GGUF, and MLX for Apple silicon.
Available precisions include BF16 across all three backends, load-time FP8 weight quantization on vLLM, GGUF Q8\_0 and Q4\_K\_M on llama.cpp, and 8-bit and 4-bit MLX builds.
We implement bidirectional attention separately for each runtime.
On vLLM, a plugin recomputes prompt representations in full-attention layers without the causal mask, achieving 99.5--99.8\% top-choice agreement with the reference implementation.
For llama.cpp, GGUF conversion marks Instant models as non-causal, while a small server patch processes each prompt as a single unit without modifying the model computation code.
On MLX, our backend implements the bidirectional prompt mask, with BF16 builds achieving 99.4--99.9\% top-choice agreement with vLLM.
Top-choice agreement does not imply numerical agreement between predicted probabilities.

\subsection{GPU Serving Performance}\label{sec:vllm-serving}
\paragraph{Single-Readout Latency on vLLM.}
On a single NVIDIA H200, BF16 Instant readouts take 24--52\,ms for 256-token prompts and 31--98\,ms for 1,024-token prompts (Table~\ref{tab:vllm-readout-latency}).
The latter setting covers most DI request lengths.
FP8 achieves comparable latency for short prompts while reducing readout latency by approximately 25--30\% at 8,192 tokens for the 9B and 27B Instant models.
Across the 4B--27B models in BF16 and FP8, Instant readouts have approximately 2--8\% higher latency than causal AutoThink readouts at 1,024 tokens and 6--14\% at 8,192 tokens.

\paragraph{Matched Model Comparison on vLLM.}
Table~\ref{tab:vllm-model-comparison} compares per-request latency with third-party decision models on 186 of 300 requests that all systems process in one forward pass.
After prompt-length weighting, this subset represents approximately 91\% of the DI request distribution.
At 4B and 9B, Instant achieves median latency within 3\,ms of the fastest comparable-scale reference, and within approximately 6\,ms at 27B, while attaining substantially higher DI 0.3 scores.
Mean latencies remain comparable across the evaluated 4B--27B models.

\begin{table}[!t]
\centering
\vspace{-3mm}
\caption{\textbf{Instant achieves higher DI 0.3 scores with competitive latency among comparable-scale decision models.} Per-request latency in milliseconds, pooled over three runs on 186 of 300 requests that all systems handle in a single forward pass. All systems receive token IDs from the same client, whereas Table~\ref{tab:vllm-readout-latency} includes tokenization. Refused denotes the fraction of rejected requests. DI 0.3 scores come from Table~\ref{tab:decision-index-results} for StartLux-Decision and the public leaderboard for reference models. $^\ddagger$Indicates our evaluation using the model's own prompt and readout, as for Decision~3.0. FlyMyJev-4B has no DI score because its 26-option limit prevents complete evaluation.}\label{tab:vllm-model-comparison}
\vspace{-2mm}
{\small\setlength{\tabcolsep}{5pt}\renewcommand{\arraystretch}{1.1}
\noindent\begin{adjustbox}{max width=\linewidth}
\begin{tabular}{lrrrrr}
\toprule
\textbf{System} & \textbf{Median} & \textbf{Mean} & {\begin{tabular}[b]{@{}r@{}}\textbf{80th}\\\textbf{percentile}\end{tabular}} & \textbf{Refused} & {\begin{tabular}[b]{@{}r@{}}\textbf{DI}\\\textbf{0.3}\end{tabular}} \\
\midrule
\mdgrouprow \multicolumn{6}{l}{\textbf{27B}} \\
\mdourrow \textbf{StartLux-Decision-27B-Instant} & 51.9 & 138.9 & 88.9 & 0 & 72.83 \\
\mdbodyrow Kev 27B & 46.1 & 126.8 & 72.7 & 0 & 56.69 \\
\mdbodyrow JEV-27B (AutoTrust AI Lab) & 46.8 & 123.1 & 76.6 & 0 & 53.64 \\
\mdbodyrow Decision 2.0 Vega (vLLM Semantic Router) & 51.5 & 130.3 & 75.7 & 0 & 56.97 \\
\mdbodyrow Perplexity Decider v1.1 & 52.4 & 143.4 & 84.2 & 0.08\% & 62.25 \\
\midrule
\mdgrouprow \multicolumn{6}{l}{\textbf{9B}} \\
\mdourrow \textbf{StartLux-Decision-9B-Instant} & 27.8 & 55.4 & 41.5 & 0 & 69.96 \\
\mdbodyrow JEV-9B (AutoTrust AI Lab) & 25.2 & 53.5 & 39.6 & 0 & 43.28 \\
\mdbodyrow Decision 2.0 Lux (vLLM Semantic Router) & 25.5 & 55.6 & 38.8 & 0 & 46.76 \\
\midrule
\mdgrouprow \multicolumn{6}{l}{\textbf{4B}} \\
\mdourrow \textbf{StartLux-Decision-4B-Instant} & 27.4 & 45.9 & 41.6 & 0 & 67.88 \\
\mdbodyrow FlyMyJev-4B (FlyMy.AI) & 24.8 & 43.6 & 38.6 & 0.08\% & -- \\
\mdbodyrow Decision 2.0 Nox (vLLM Semantic Router) & 25.2 & 46.7 & 38.7 & 0 & 44.21 \\
\mdbodyrow H2O-Lightning-4B (H2O.ai) & 27.4 & 45.3 & 42.3 & 0 & 51.43$^\ddagger$ \\
\midrule
\mdgrouprow \multicolumn{6}{l}{\textbf{2B and 3B}} \\
\mdourrow \textbf{StartLux-Decision-2B-Instant} & 24.6 & 35.1 & 36.8 & 0 & 59.14 \\
\mdbodyrow d1-3B (Liquid AI) & 16.7 & 26.6 & 24.1 & 0 & 48.99 \\
\mdbodyrow Decision 2.0 Sol (vLLM Semantic Router) & 22.8 & 33.3 & 33.1 & 0 & 30.05 \\
\midrule
\mdgrouprow \multicolumn{6}{l}{\textbf{0.6B to 1.7B}} \\
\mdourrow \textbf{StartLux-Decision-0.8B-Instant} & 24.8 & 33.7 & 36.6 & 0 & 53.75 \\
\mdbodyrow Decision 2.0 Kai, 0.6B (vLLM Semantic Router) & 10.2 & 17.7 & 19.6 & 0.08\% & 16.68 \\
\mdbodyrow Bosun 3.1 1.7B (Hanno Labs) & 11.8 & 20.8 & 22.6 & 0 & 20.55 \\
\mdbodyrow Decision 2.0 Eos, 0.8B (vLLM Semantic Router) & 24.5 & 32.6 & 34.1 & 0 & 20.57 \\
\bottomrule
\end{tabular}
\end{adjustbox}\par}
\end{table}

\paragraph{GGUF Inference on llama.cpp.}
We evaluate GGUF Q8\_0 builds on an H200 using the same request subset and prompt-length weighting as the vLLM comparison (Table~\ref{tab:llamacpp-model-comparison}); reference models are converted using the llama.cpp converter.
The 4B Instant model achieves the lowest median latency among comparable-scale references, while the 9B model is within 3\,ms of the fastest.
At 2B and 27B, median latency is approximately 5\,ms and 14\,ms higher than d1-3B and Kev 27B, respectively.
At 0.8B, Instant matches Bosun 3.1 1.7B in median latency, with broadly comparable mean latencies across similar scales.
Instant GGUF inference uses standard llama.cpp model computation with the server-side prompt handling described in Section~\ref{sec:inference-engines}.

\begin{table}[!t]
\centering
\vspace{-3mm}
\caption{\textbf{GGUF Q8\_0 Instant models achieve competitive single-pass request latency on llama.cpp.} Per-request latency in milliseconds on one H200 using llama.cpp b11549, pooled over three runs with the requests, weighting, and client of Table~\ref{tab:vllm-model-comparison}. The server processes one prompt in a micro-batch of up to 16,384 tokens to support bidirectional attention; this configuration increases causal-model median latency by 1--3\%. For models with a separate prediction head, timings cover the backbone and exclude the head's small matrix product. Refused denotes the fraction of rejected requests.}\label{tab:llamacpp-model-comparison}
\vspace{-2mm}
{\small\setlength{\tabcolsep}{5pt}\renewcommand{\arraystretch}{1.1}
\noindent\begin{adjustbox}{max width=\linewidth}
\begin{tabular}{lrrrr}
\toprule
\textbf{System} & \textbf{Median} & \textbf{Mean} & {\begin{tabular}[b]{@{}r@{}}\textbf{80th}\\\textbf{percentile}\end{tabular}} & \textbf{Refused} \\
\midrule
\mdgrouprow \multicolumn{5}{l}{\textbf{27B}} \\
\mdourrow \textbf{StartLux-Decision-27B-Instant} & 140.7 & 519.4 & 301.9 & 0 \\
\mdbodyrow Kev 27B & 126.6 & 451.3 & 250.3 & 0 \\
\mdbodyrow JEV-27B (AutoTrust AI Lab) & 138.8 & 489.0 & 261.4 & 0 \\
\mdbodyrow Decision 2.0 Vega (vLLM Semantic Router) & 161.4 & 509.3 & 261.1 & 0 \\
\mdbodyrow Perplexity Decider v1.1 & 143.6 & 537.8 & 292.8 & 0.08\% \\
\midrule
\mdgrouprow \multicolumn{5}{l}{\textbf{9B}} \\
\mdourrow \textbf{StartLux-Decision-9B-Instant} & 53.8 & 185.6 & 106.9 & 0 \\
\mdbodyrow JEV-9B (AutoTrust AI Lab) & 50.9 & 175.3 & 91.8 & 0 \\
\mdbodyrow Decision 2.0 Lux (vLLM Semantic Router) & 55.6 & 181.4 & 92.4 & 0 \\
\midrule
\mdgrouprow \multicolumn{5}{l}{\textbf{4B}} \\
\mdourrow \textbf{StartLux-Decision-4B-Instant} & 40.6 & 135.2 & 75.6 & 0 \\
\mdbodyrow FlyMyJev-4B (FlyMy.AI) & 42.0 & 139.8 & 69.3 & 0.08\% \\
\mdbodyrow Decision 2.0 Nox (vLLM Semantic Router) & 41.0 & 130.7 & 63.7 & 0 \\
\mdbodyrow H2O-Lightning-4B (H2O.ai) & 41.9 & 133.4 & 66.6 & 0 \\
\midrule
\mdgrouprow \multicolumn{5}{l}{\textbf{2B and 3B}} \\
\mdourrow \textbf{StartLux-Decision-2B-Instant} & 25.7 & 83.2 & 44.9 & 0 \\
\mdbodyrow d1-3B (Liquid AI) & 20.2 & 72.6 & 35.8 & 0 \\
\mdbodyrow Decision 2.0 Sol (vLLM Semantic Router) & 26.2 & 80.1 & 39.7 & 0 \\
\midrule
\mdgrouprow \multicolumn{5}{l}{\textbf{0.6B to 1.7B}} \\
\mdourrow \textbf{StartLux-Decision-0.8B-Instant} & 21.1 & 68.1 & 36.1 & 0 \\
\mdbodyrow Decision 2.0 Kai, 0.6B (vLLM Semantic Router) & 15.0 & 45.4 & 21.1 & 0.08\% \\
\mdbodyrow Bosun 3.1 1.7B (Hanno Labs) & 21.1 & 63.9 & 30.6 & 0 \\
\mdbodyrow Decision 2.0 Eos, 0.8B (vLLM Semantic Router) & 21.6 & 66.0 & 31.3 & 0 \\
\bottomrule
\end{tabular}
\end{adjustbox}\par}
\end{table}

\subsection{Local Inference on Apple Silicon}\label{sec:llamacpp-mlx-serving}
We evaluate local inference using MLX on an Apple M5 Pro with 64\,GB of unified memory (Table~\ref{tab:mlx-readout-latency}).
For 1,024-token prompts, BF16 Instant readout latency ranges from 59\,ms at 0.8B to approximately 1.1\,s at 9B.
While the 27B models exceed the available memory in BF16, 8-bit and 4-bit quantization enable their local deployment.

\begin{table}[!t]
\centering
\vspace{-3mm}
\caption{\textbf{MLX supports local BF16 and quantized inference across model sizes.} Median single-readout latency in milliseconds at the specified prompt length, measured using mlx-lm on an Apple M5 Pro with 64\,GB of unified memory. Measurements use three runs after warm-up, with one request in flight. A dash indicates unavailable BF16 results for the 27B models due to memory constraints. BF16 baseline rows are shaded.}\label{tab:mlx-readout-latency}
\vspace{-2mm}
{\small\setlength{\tabcolsep}{6pt}\renewcommand{\arraystretch}{1.1}
\noindent\begin{tabular}{lrrrrrrrrr}
\toprule
\textbf{Precision} & \textbf{Prompt length} & \multicolumn{5}{c}{\textbf{Instant}} & \multicolumn{3}{c}{\textbf{AutoThink}} \\
\cmidrule(lr){3-7}\cmidrule(lr){8-10}
 & \textbf{(tokens)} & \textbf{0.8B} & \textbf{2B} & \textbf{4B} & \textbf{9B} & \textbf{27B} & \textbf{4B} & \textbf{9B} & \textbf{27B} \\
\midrule
\mdourrow \textbf{BF16} & 256 & 17 & 34 & 83 & 166 & -- & 86 & 158 & -- \\
\mdourrow  & 1,024 & 59 & 139 & 522 & 1,087 & -- & 587 & 965 & -- \\
\mdourrow  & 4,096 & 277 & 844 & 2,250 & 4,461 & -- & 2,257 & 4,116 & -- \\
\mdourrow  & 8,192 & 752 & 1,942 & 5,085 & 9,836 & -- & 4,800 & 8,752 & -- \\
\midrule
\mdbodyrow 8-bit & 256 & 26 & 36 & 95 & 186 & 1,384 & 94 & 171 & 974 \\
\mdbodyrow  & 1,024 & 66 & 152 & 602 & 1,124 & 4,294 & 466 & 1,087 & 3,839 \\
\mdbodyrow  & 4,096 & 535 & 876 & 2,295 & 4,569 & 17,008 & 2,339 & 4,404 & 15,824 \\
\mdbodyrow  & 8,192 & 1,080 & 2,002 & 5,815 & 9,850 & 32,002 & 4,875 & 9,053 & 38,069 \\
\midrule
\mdbodyrow 4-bit & 256 & 26 & 35 & 89 & 177 & 946 & 88 & 169 & 711 \\
\mdbodyrow  & 1,024 & 64 & 144 & 571 & 1,041 & 3,308 & 527 & 993 & 3,023 \\
\mdbodyrow  & 4,096 & 488 & 851 & 2,547 & 4,276 & 13,993 & 2,275 & 4,102 & 12,496 \\
\mdbodyrow  & 8,192 & 1,067 & 1,992 & 5,571 & 9,374 & 28,697 & 4,730 & 8,435 & 33,465 \\
\bottomrule
\end{tabular}\par}
\end{table}

\subsection{Quantized Inference}\label{sec:quantized-inference}
StartLux-Decision supports reduced-precision inference with FP8 on vLLM, GGUF Q8\_0 and Q4\_K\_M on llama.cpp, and 8-bit and 4-bit builds on MLX.
FP8 and Q8\_0 largely preserve BF16 decision performance across the evaluated settings, whereas Q4\_K\_M can incur larger losses, reaching 4.3 points for the 27B AutoThink model on DecisionPivot (Table~\ref{tab:quantization-results}).
MLX 8-bit builds achieve 99.5--99.6\% top-choice agreement with the vLLM BF16 reference, compared with 96.3--97.5\% for 4-bit builds.

\begin{table}[!t]
\centering
\caption{\textbf{FP8 and GGUF Q8\_0 largely preserve decision performance, while Q4\_K\_M can incur larger losses.} Results report DI 0.3 public scores and DecisionPivot mean chance-corrected accuracy across seven scoring layers on 7,800 core examples; the latter differs from pair accuracy in Section~\ref{sec:decisionpivot}. BF16 and FP8 use vLLM, while GGUF formats use llama.cpp b11549, so cross-format comparisons also reflect backend differences. BF16 DI 0.3 runs are separate from Table~\ref{tab:decision-index-results}; a dash denotes an unavailable result.}\label{tab:quantization-results}
{\small\setlength{\tabcolsep}{5pt}\renewcommand{\arraystretch}{1.1}
\noindent\begin{adjustbox}{max width=\linewidth}
\begin{tabular}{lrrrrrr}
\toprule
\textbf{Model} & \multicolumn{2}{c}{\textbf{DI 0.3}} & \multicolumn{4}{c}{\textbf{DecisionPivot}} \\
\cmidrule(lr){2-3}\cmidrule(lr){4-7}
 & \textbf{BF16} & \textbf{FP8} & \textbf{BF16} & \textbf{FP8} & \textbf{Q8\_0} & \textbf{Q4\_K\_M} \\
\midrule
\mdgrouprow \multicolumn{7}{l}{\textbf{Instant}} \\
\mdbodyrow 0.8B & 53.81 & 53.24 & 9.5 & 8.7 & 9.7 & 8.7 \\
\mdbodyrow 2B & 59.18 & 59.16 & 23.5 & 23.6 & 23.7 & 21.9 \\
\mdbodyrow 4B & 67.84 & 67.87 & 47.5 & 46.2 & 46.2 & 43.7 \\
\mdbodyrow 9B & 69.98 & 69.84 & 53.2 & 52.8 & 53.2 & 50.6 \\
\mdbodyrow 27B & 72.89 & 72.91 & 80.5 & 80.5 & 80.7 & 80.2 \\
\midrule
\mdgrouprow \multicolumn{7}{l}{\textbf{AutoThink}} \\
\mdbodyrow 4B & 68.11 & 68.09 & 65.3 & 65.6 & 65.6 & 63.5 \\
\mdbodyrow 9B & 70.47 & 70.19 & 63.8 & 66.4 & 64.7 & 64.2 \\
\mdbodyrow 27B & 73.52 & 73.20 & 88.6 & 87.3 & 89.5 & 84.3 \\
\bottomrule
\end{tabular}
\end{adjustbox}\par}
\end{table}

\FloatBarrier

\section{Conclusion}\label{sec:conclusion}
StartLux-Decision combines efficient single-pass prediction with confidence-guided reasoning through a unified probability interface.
\rlcdtwo{} improves reasoning termination and efficiency through proper-score rewards and a soft length penalty, while largely preserving held-out decision accuracy.
Evaluations on Decision Index and DecisionPivot demonstrate strong decision performance, while policy-aware calibration mitigates reliability distortions induced by selection and routing.
Support for vLLM, llama.cpp, and MLX enables deployment across GPU and local inference environments.
Future work will focus on generalization beyond represented task families and robust probability reliability under distribution shift.
Together, these results advance efficient and reliable decision modeling by integrating direct prediction, selective reasoning, and calibrated confidence.

\FloatBarrier
\mdBibliography
\clearpage
\appendix
\counterwithin{table}{section}
\section{Data Sources}\label{sec:A}
This appendix documents the training mixtures, label provenance, and data integrity checks for the pipeline in Section~\ref{sec:model-training}.

\subsection{Record Format}\label{sec:A.1}
Each training record represents a typed decision about an input state, specifying candidate options, target probabilities, label provenance, and a parent-group identifier.
Records derived from the same state share a parent group, enabling group-level splitting and decontamination.
Overlength records are excluded rather than truncated, and a record-level hash deterministically permutes option order to reduce positional shortcuts.

\subsection{Data Composition Across Training Stages}\label{sec:A.2}
\emph{Lux-Release}, used in Stage 1, combines public training splits, community decision corpora, programmatically generated examples, knowledge and commonsense multiple-choice questions, and smaller collections of tool-use, safety, and challenging decisions.
Table~\ref{tab:A.1} reports its composition and label provenance, covering approximately 2.7 million records for the 0.8B--9B models and 0.8 million for 27B.

\begin{table}[!htbp]
\centering
\caption{\textbf{Composition and label provenance of Lux-Release across model scales.} The 27B mixture places greater emphasis on benchmark training splits and knowledge-oriented multiple-choice questions. Label types: H = human annotation or source gold label; T = model-generated or model-assigned label; P = programmatically computed label.}\label{tab:A.1}
{\small\setlength{\tabcolsep}{5pt}\renewcommand{\arraystretch}{1.1}
\noindent\begin{adjustbox}{max width=\linewidth}
\begin{tabular}{>{\raggedright\arraybackslash}p{\mdw{0.56}}>{\raggedleft\arraybackslash}p{\mdw{0.14}}>{\raggedleft\arraybackslash}p{\mdw{0.14}}>{\raggedright\arraybackslash}p{\mdw{0.16}}}
\toprule
\textbf{Source family} & \multicolumn{2}{c}{\textbf{Share of records}} & \textbf{Labels} \\
\cmidrule(lr){2-3}
 & \textbf{0.8B--9B} & \textbf{27B} & \\
\midrule
\mdbodyrow Public training splits: classification, inference, question answering, safety, preference, and tool use & 43.6\% & 18.6\% & H, P, T \\
\mdbodyrow Community corpora of typed decisions & 24.4\% & 17.2\% & T, P, H \\
\mdbodyrow Training and development splits of 15 DI benchmarks & 10.0\% & 26.4\% & H \\
\mdbodyrow Knowledge and commonsense multiple-choice questions & 7.1\% & 22.3\% & H \\
\mdbodyrow Programmatically generated questions & 5.8\% & 2.5\% & P \\
\mdbodyrow Small challenging decision datasets & 0.9\% & 1.4\% & T, P \\
\mdbodyrow Tool use, our in-house pool, safety, and palettes & 8.2\% & 11.5\% & H, source labels \\
\midrule
\mdgrouprow \textbf{Total records} & $\sim$2.7M & $\sim$0.8M & -- \\
\bottomrule
\end{tabular}
\end{adjustbox}\par}
\end{table}

\emph{Lux-Rounds}, used in Stage 2, contains approximately 1.7 million records primarily from benchmark training splits, related public datasets, and additional decision families.
Other sources include programmatically generated and model-written examples, select-all questions, teacher-labeled decisions, a subset of Lux-Release, and reconstructed GSM8K training problems.
\emph{Lux-Think}, used in Stage 3, contains approximately 4,000--21,000 supervised reasoning examples per model (Section~\ref{sec:model-training}).

\subsection{Benchmark-Specific Training Coverage}\label{sec:A.3}
Table~\ref{tab:A.2} summarizes task-related Stage 2 training sources for Decision Index 0.2.1 benchmarks in the preview-model mixture.
T marks the official training split of a benchmark, which is separate from its evaluation items, and no evaluation item is used in training for either the preview or the released models.

\begin{table}[!htbp]
\centering
\caption{\textbf{Benchmark-specific training coverage in the Stage 2 mixture of the preview models.} T = official training split of the benchmark or its source; G = independently generated or model-written data; R = related public data; -- = no targeted Stage 2 data. This earlier mixture is not an exhaustive record of the final released checkpoints.}\label{tab:A.2}
{\small\setlength{\tabcolsep}{5pt}\renewcommand{\arraystretch}{1.1}
\noindent\begin{adjustbox}{max width=\linewidth}
\begin{tabular}{>{\raggedright\arraybackslash}p{\mdw{0.78}}>{\raggedright\arraybackslash}p{\mdw{0.22}}}
\toprule
\textbf{Benchmarks} & \textbf{Stage 2 sources} \\
\midrule
\mdbodyrow BANKING77, CLINC150, HoVer, ContractNLI, ANLI, VAST, NLI4CT, RAGTruth, iSarcasmEval, GSM8K, New Yorker, WinoGrande, HellaSwag, cfcolor, Amazon ESCI, ACOS, When2Call & T \\
\mdbodyrow API-Bank & R \\
\mdbodyrow BFCL & R \\
\mdbodyrow GPQA Diamond, MMLU-Pro & R (layout only) \\
\mdbodyrow Habermas Machine, ForecastBench, Humicroedit & T \\
\mdbodyrow appliance simulator, PhishNChips, CLadder, MuSR, ChessBench, POP909-CL, BPoMP, BRIGHT, BBH & G \\
\mdbodyrow SATA-Bench & G, T (upstream sources) \\
\mdbodyrow FinEntity & R \\
\mdbodyrow ToolRet, HLE, CRUXEval & -- \\
\bottomrule
\end{tabular}
\end{adjustbox}\par}
\end{table}

\subsection{Licensing and Data Usage}\label{sec:A.4}
The inference code is released under Apache-2.0, while model weights are distributed under CC BY-NC 4.0 for non-commercial use, with separate licensing available for commercial applications.
Training sources have varying licensing and usage conditions, including public datasets and commercially accessed model-generated examples.
Model-generated examples are not redistributed.

\subsection{Generated and Model-Written Data}\label{sec:A.5}
Programmatically generated examples pair independent templates with deterministically computed labels.
No evaluation items are supplied to the generators, and the procedural templates are independent of those used in DecisionPivot.
Model-written examples are retained only when two verifier models independently recover the proposed label without seeing the writer's answer.

\subsection{Data Integrity Checks}\label{sec:A.6}
\paragraph{Data Decontamination.}
We screen training records against Decision Index, public JevBench items, DecisionPivot, and other held-out sets, excluding related records by parent group to reduce training--evaluation overlap.

\paragraph{Shortcut Detection.}
We audit dataset sources for answer-position and option-format shortcuts using content-blind baselines, such as selecting a fixed position or the longest option without reading the input state.
These audits identify shortcuts in the original GSM8K evaluation format, motivating the reconstructed questions in DI 0.3.

\section{Training Implementation and Additional Analyses}\label{sec:B}
Stage 1 and \rlcdtwo{} optimize all language-model parameters, whereas Stage 2 and reasoning supervision use adapters subsequently merged into the model weights.
The pretrained vision tower remains frozen throughout training (Section~\ref{sec:model-design}).

\subsection{Bidirectional Prompt Attention}\label{sec:B.1}
Bidirectional prompt attention is enabled by removing the causal mask from full-attention layers during single-pass prediction, while linear-attention layers retain causal recurrence.
With bidirectional attention disabled, the modified implementation produces outputs bit-identical to those of the original causal implementation.
Runtime integration and cross-backend validation are described in Section~\ref{sec:inference-engines}.

\subsection{Additional Training Ablations}\label{sec:B.2}
Table~\ref{tab:B.1} summarizes additional ablations of probability optimization, parameter updates, reasoning termination, and \rlcdtwo{} initialization.

\begin{table}[!t]
\centering
\caption{\textbf{Targeted ablations of the StartLux-Decision training design.} Results use the 4B model unless otherwise specified.}\label{tab:B.1}
{\small\setlength{\tabcolsep}{5pt}\renewcommand{\arraystretch}{1.1}
\noindent\begin{adjustbox}{max width=\linewidth}
\begin{tabular}{>{\raggedright\arraybackslash}p{\mdw{0.25}}>{\raggedright\arraybackslash}p{\mdw{0.39}}>{\raggedright\arraybackslash}p{\mdw{0.36}}}
\toprule
\textbf{Design aspect} & \textbf{Alternative configuration} & \textbf{Observation} \\
\midrule
\mdbodyrow Direct probability optimization & Sampled policy-gradient optimization of the single-pass objective in Appendix~\ref{sec:direct-probability-optimization} & Calibration error of 5.3 points, versus 3.9 with exact gradients \\
\mdbodyrow Parameter-update strategy & Train Stages 1 and 2 using a single adapter on the unmodified base model & DI 0.2.1 scores of 63.3 and 58.2, versus 64.8 with full-parameter training \\
\mdbodyrow Reasoning termination & Increase the generation limit fourfold for a model exhibiting repetitive reasoning loops & Termination increases from 11\% to 13\%, with unchanged accuracy \\
\mdbodyrow \rlcdtwo{} initialization & Apply policy-gradient training to a checkpoint that frequently produces empty reasoning traces & Reasoning is skipped on approximately two-thirds of decisions; 57 reasoning tokens per answer on average \\
\bottomrule
\end{tabular}
\end{adjustbox}\par}
\end{table}

\subsection{Calibration and Resolution under Proper Scoring Rules}\label{sec:brier-decomposition}
The Brier decomposition separates the expected score into calibration, resolution, and outcome uncertainty:
\begin{equation}
\label{eq:brier-decomposition}
\mathbb{E}\, S_{\mathrm{B}} = -\,\mathbb{E}\, \big\| p - \bar t(p) \big\|^{2} + \mathbb{E}\,
\big\| \bar t(p) - \bar t \big\|^{2} - \mathbb{E}\, \big\| t - \bar t \big\|^{2},
\end{equation}
where $\bar t(p)=\mathbb{E}[t\mid p]$ is the conditional mean target and $\bar t=\mathbb{E}[t]$ is the overall mean target.
The terms represent negative miscalibration, positive resolution, and negative outcome uncertainty, respectively.
For a fixed evaluation distribution, outcome uncertainty is independent of the model.
A higher expected score can therefore reflect improved calibration, resolution, or both, without changing the predicted option.

\subsection{Direct Probability Optimization}\label{sec:direct-probability-optimization}
We separately evaluate direct optimization of single-pass decision probabilities, independent of \rlcdtwo{} and the training procedure for the released Instant models.
Unlike sampled policy-gradient optimization, this objective permits exact differentiation with respect to model parameters.
We augment the logarithmic score with decision-utility and record-level terms, regularized toward a reference model:
\begin{equation}
\label{eq:direct-probability-objective}
\mathcal{L}=-S_{\mathrm{L}}(p_\theta,t)-\frac{1}{2}p_\theta^{\top}Ut-\frac{1}{2}\prod_{j\in g}p_{\theta,j}^{\top}t_j+0.05\,\mathrm{KL}\big(p_\theta\,\|\,p_{\mathrm{ref}}\big).
\end{equation}
Here $U_{ky}$ denotes the utility of predicting option $k$ when option $y$ is correct, assigning unit utility to exact matches and allowing partial credit for ordered levels.
The product extends over questions $j$ within record $g$, rewarding joint correctness.
The utility terms can encourage sharper predictions but are not proper scoring rules in isolation; strict propriety is therefore not guaranteed for the combined objective.

In a separate 4B single-pass experiment using the same training data as cross-entropy, this objective reduces calibration error from 5.1 to 3.9 points (approximately 23\%) and Brier loss from 0.278 to 0.271.
DI 0.2.1 scores remain comparable, at 52.49 for the baseline and 52.57 for the optimized objective.
Sampled-answer policy gradients yield a similar DI score of 52.55 but no calibration improvement, favoring exact gradients for calibration in this setting.

The optimized model produces sharper raw predictions, with the fitted temperature increasing from 1.37 to 1.60.
Since the contributions of individual objective terms are not isolated, the calibration improvement is attributed to their combined effect.

\section{Policy-Aware Calibration: Proofs and Details}\label{sec:C}
We formalize selection- and routing-induced calibration effects and provide evaluation protocols and threshold-certification results.
Experiments use publicly available model weights, except for the external actor-to-reasoner analysis based on public system records.

\subsection{Setting and Notation}\label{sec:C.1}
Let $x$ be an input, $y$ its correct option, and $p_0$ and $p_1$ the single-pass and post-reasoning distributions. Define confidence $c(p)=\max_k p_k$ and prediction $a(p)=\arg\max_k p_k$. A think-gate routes an input to reasoning if $c(p_0)<t$. Marginal calibration concerns individual predictions; policy-aware calibration concerns predictions returned after selection or routing.

For best-of-$N$, let $\theta$ be latent problem easiness with density $h(\theta)$. Conditional on $\theta$, the candidate correctness--score pairs $(Y_i,S_i)$ are independent and identically distributed, with score density $f(s\mid\theta)$ and CDF $F(s\mid\theta)$. Let $m(s\mid\theta)=\Pr(Y_i=1\mid S_i=s,\theta)$ and $m(s)=\Pr(Y_i=1\mid S_i=s)$. Assuming continuous scores to avoid ties, the policy returns $I^*=\arg\max_i S_i$.

\subsection{Theoretical Properties}\label{sec:C.2}
\paragraph{Proposition 1 (Selection by the reported score).}
Suppose (i) $m(s\mid\theta)$ is non-decreasing in $\theta$, and (ii) $F(s\mid\theta)$ is non-increasing in $\theta$ for each score $s$. Then $a_N(s)=\Pr(Y_{I^*}=1\mid S_{I^*}=s)$ is non-increasing in $N$, with $a_1(s)=m(s)$. Thus, a larger candidate pool cannot improve accuracy at a fixed score. If the individual-response verifier is calibrated, $m(s)=s$, selected responses are weakly overconfident; strict overconfidence requires stronger conditions.

\emph{Proof.} Conditional on $\theta$, the density of the maximum score is $Nf(s\mid\theta)F(s\mid\theta)^{N-1}$. Conditional independence gives
\[
a_N(s)=\int m(s\mid\theta)w_N(\theta;s)\,d\theta,\qquad w_N(\theta;s)\propto h(\theta)f(s\mid\theta)F(s\mid\theta)^{N-1}.
\]
Increasing $N$ by one multiplies the posterior weight by $F(s\mid\theta)$, non-increasing in $\theta$. This downward shift in monotone likelihood-ratio order cannot increase the expectation of non-decreasing $m(s\mid\theta)$.\nobreak\hspace{0.4em}$\blacksquare$

Using the verifier's mean score on competing candidates as a proxy for easiness, responses scored 0.8--0.9 by the 9B verifier achieve 75.6\%, 93.4\%, and 98.3\% accuracy in low-, medium-, and high-easiness strata. This pattern is consistent with, but does not establish, the first assumption.

\paragraph{Proposition 2 (Routing by an additional score).}
Let $G=\{c(p_0)<t\}$, $Q=c(p_1)$, and $Z=\mathbf{1}\{a(p_1)=y\}$. If $\Pr(Z=1\mid Q=q,c(p_0)=u)$ is non-decreasing in $u$, then
\[
\Pr(Z=1\mid Q=q,G)\leq\Pr(Z=1\mid Q=q)\leq\Pr(Z=1\mid Q=q,G^c).
\]
The inequalities reverse for non-increasing dependence. If $p_1$ is marginally calibrated and both routes have positive probability at $q$, calibration within the reasoning route is equivalent to $Z\perp G\mid Q=q$.

\emph{Proof.} At fixed $Q=q$, route-specific accuracies average the same monotone function over initial confidences below or above $t$, giving the inequalities. Marginal calibration makes $q$ a weighted average of both route accuracies. If one route's accuracy equals $q$, the other must also equal $q$, which for binary $Z$ and $G$ is equivalent to conditional independence.\nobreak\hspace{0.4em}$\blacksquare$

We diagnose residual routing information using the AUROC of initial confidence for predicting correctness within bins of post-reasoning confidence. Values above or below 0.5 indicate residual ranking direction, but AUROC 0.5 does not establish conditional independence. The analysis also applies to cascades.

\paragraph{Proposition 3 (Marginal certification does not imply route-wise certification).}
Let $C$ be the confidence returned by a gated system, which accepts predictions when $C\geq\lambda$. Define $e=\Pr(\mathrm{wrong}\mid C\geq\lambda)$, $e_r=\Pr(\mathrm{wrong}\mid C\geq\lambda,r)$, and $w_r=\Pr(r\mid C\geq\lambda)$. Then
\[
e=w_{\mathrm{plain}}e_{\mathrm{plain}}+w_{\mathrm{thought}}e_{\mathrm{thought}}.
\]
For $w_{\mathrm{thought}}>0$, $e\leq\alpha$ implies only
\[
e_{\mathrm{thought}}\leq\frac{\alpha-w_{\mathrm{plain}}e_{\mathrm{plain}}}{w_{\mathrm{thought}}},
\]
which can exceed $\alpha$ when $e_{\mathrm{plain}}<\alpha$. Conversely, certifying $e_r\leq\alpha$ on every route with positive acceptance probability guarantees the same bound $e\leq\alpha$ for the accepted predictions overall.

\emph{Proof.} Apply the law of total probability conditional on acceptance.\nobreak\hspace{0.4em}$\blacksquare$

\paragraph{Proposition 4 (Calibration conditional on policy statistics).}
Let $T$ contain the information determining whether a candidate is returned, and let $q$ be its reported confidence. If $\Pr(Z=1\mid q,T)=q$, where $Z$ indicates correctness, then any return event $R$ determined by $(q,T)$ preserves calibration:
\[
\Pr(Z=1\mid q,R)=\mathbb{E}[\Pr(Z=1\mid q,T)\mid q,R]=q.
\]
\emph{Proof.} The equality follows from the tower property and the measurability of $R$ with respect to $(q,T)$.\nobreak\hspace{0.4em}$\blacksquare$

Policy-aware maps approximate conditional correctness using candidate scores, candidate counts, or routing confidence (Section~\ref{sec:policy-aware-calibration}); finite-sample estimation does not guarantee exact conditional calibration.

\paragraph{Proposition 5 (Optimal gating under scoring and computation costs).}
Let $S$ be a proper scoring rule, $g(x)\in[0,1]$ the probability of reasoning, and $c(\tau)$ the cost of a reasoning trace. For a fixed conditional reasoning policy, define
\[
J=\mathbb{E}\left[(1-g(x))S(p_0(x),y)+g(x)(S(p_1(x,\tau),y)-c(\tau))\right].
\]
Writing
\[
v(x)=\mathbb{E}[S(p_1(x,\tau),y)-c(\tau)\mid x]-\mathbb{E}[S(p_0(x),y)\mid x],
\]
the objective is affine in $g(x)$ and is maximized pointwise by $g^*(x)=\mathbf{1}\{v(x)>0\}$, with either choice optimal at $v(x)=0$. For a fixed gate and strictly proper $S$, the optimal post-reasoning readout given statistic $\phi(x,\tau)$ is the conditional distribution of $y$ given $\phi$ under the gate-weighted distribution.

\emph{Proof.} The coefficient of $g(x)$ in the conditional objective is $v(x)$, giving the optimal gate. Strict propriety yields the optimal conditional readout.\nobreak\hspace{0.4em}$\blacksquare$

This characterization does not establish optimality of the released models' fixed confidence threshold.

\subsection{Evaluation Protocols}\label{sec:C.3}
Best-of-$N$ experiments use nine verifiers (three StartLux-Decision models, two third-party decision models, and four chat models) on held-out variants of GSM8K training problems, five unseen arithmetic datasets, and reasoning tasks. Routing experiments use twelve models on diagnostic decisions spanning 39 task families, reasoning decisions, and CJ-Bench, a collection of challenging decisions in Chinese.

Signed overconfidence is mean reported confidence minus empirical accuracy, in percentage points. The route effect is the difference between routed and non-routed accuracy at matched reported confidence. Acceptance thresholds use a 95\% fixed-sequence Clopper--Pearson procedure on the calibration half of each of 20 source-problem splits.

\subsection{Results Across Decision Pipelines}\label{sec:C.4}
Table~\ref{tab:C.1} reports calibration gaps across four decision pipelines and an external actor-to-reasoner system.

\begin{table}[!htbp]
\centering
\caption{\textbf{Policy-aware calibration reduces signed confidence--accuracy gaps across selection and routing pipelines.} Overconfidence is mean reported confidence minus empirical accuracy, in percentage points. The external actor-to-reasoner row uses publicly available system records; all other results come from our own evaluations.}\label{tab:C.1}
{\small\setlength{\tabcolsep}{5pt}\renewcommand{\arraystretch}{1.1}
\noindent\begin{adjustbox}{max width=\linewidth}
\begin{tabular}{>{\raggedright\arraybackslash}p{\mdw{0.19}}>{\raggedright\arraybackslash}p{\mdw{0.35}}>{\raggedright\arraybackslash}p{\mdw{0.25}}>{\raggedleft\arraybackslash}p{\mdw{0.21}}}
\toprule
\textbf{Pipeline} & \textbf{Settings} & \textbf{Conventional calibration} & \textbf{Policy-aware calibration} \\
\midrule
\mdbodyrow Best-of-$N$ & Unseen datasets; five verifiers; gap increases with $N$; $N=2$--$32$; fitted at $N=1,4,16$ & +0.8 to +14.6 & $-$1.2 to +0.3 \\
\mdbodyrow Think-gate & Diagnostic decisions; nine models; reasoning rates of 10--50\% & up to +16 on routed predictions & $-$1.5 to +2.1 \\
\mdbodyrow Cascade & Diagnostic decisions; seven model pairs from two teams; 30\% escalated & +2.3 to +8.4 on escalated predictions & $-$0.6 to +0.5 \\
\mdbodyrow Select-then-think & Ten settings; fitted at the two or three smallest $N$ & +2.1 to +7.3 at the largest $N$ & $-$1.4 to +0.7 at every $N$ \\
\mdbodyrow Actor $\rightarrow$ reasoner (external) & 21 settings; 30\% deferred & $-$10.7 to +18.1 on deferred predictions & $-$7.4 to +4.4 \\
\bottomrule
\end{tabular}
\end{adjustbox}\par}
\end{table}

The route effect is strongly associated with residual routing-confidence information for think-gates and cascades (Figure~\ref{fig:C.1}). An isotonic map of post-reasoning confidence within quantile strata of initial confidence reduces the route-specific gaps to within approximately 2 percentage points across nine models. Public records of System Switch~\citep{bailo2026systemswitch}, where uncertain actor decisions are deferred to a reasoner, show an 8.3-point mean absolute gap between overall reasoner accuracy and accuracy on deferred items, versus 2.1 points after route-aware calibration.

\begin{figure}[!tbp]
\renewcommand{\thefigure}{C.\arabic{figure}}
\setcounter{figure}{0}
\centering
\adjincludegraphics[max width=\linewidth,max height=0.5\textheight]{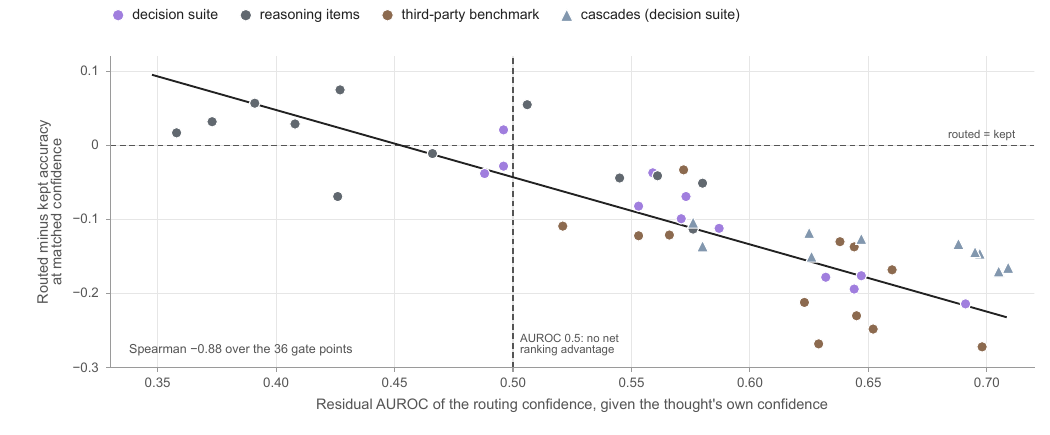}
\caption{\textbf{Residual routing information is strongly associated with route-dependent accuracy differences.} Residual AUROC measures how routing confidence ranks correctness within bins of post-reasoning confidence. The route effect is routed minus non-routed accuracy at matched confidence (Spearman $-0.88$ across 36 model--dataset settings; triangles mark cascades).}\label{fig:C.1}
\end{figure}

\subsection{Policy-Specific Threshold Certification}\label{sec:C.5}
Table~\ref{tab:C.2} compares conventional and policy-specific threshold certification. Policy-specific thresholds reduce observed error violations, although some routes accept no predictions.

\begin{table}[!htbp]
\centering
\caption{\textbf{Policy-specific threshold certification reduces violations of target error rates among accepted predictions.} Entries give accepted-prediction error rates and, in parentheses, the number of evaluation splits (out of 20) exceeding target $\alpha$. Some policy-specific thresholds accept no predictions on a route.}\label{tab:C.2}
{\small\setlength{\tabcolsep}{5pt}\renewcommand{\arraystretch}{1.1}
\noindent\begin{adjustbox}{max width=\linewidth}
\begin{tabular}{>{\raggedright\arraybackslash}p{\mdw{0.38}}>{\raggedright\arraybackslash}p{\mdw{0.31}}>{\raggedright\arraybackslash}p{\mdw{0.31}}}
\toprule
\textbf{Setting} & \textbf{Conventional thresholds} & \textbf{Policy-specific thresholds} \\
\midrule
\mdourrow StartLux-Decision 4B / 9B / 27B, best-of-32, unseen datasets, $\alpha=0.05$ & 10.05 / 9.27 / 7.57\% (19 / 20 / 20) & 3.5--4.3\% (0--4) \\
\mdbodyrow Same models; best-of-256; 0.8B generator & 12.9 / 11.6 / 7.8\% (20 each) & 3.4 / 3.3 / 3.7\% (0--1) \\
\mdbodyrow All 67 pairs with nonzero coverage; best-of-16, $\alpha=0.05$ & Target exceeded in 44 of 67 pairs & Target exceeded in 0 of 67 pairs \\
\mdourrow StartLux-Decision-4B, think-gate, CJ-Bench, $\alpha=0.05$, 10\% / 20\% think rate & Reasoning-route error 27.9 / 27.3\% (20 / 20) & No reasoning-route predictions accepted; system error 4.6\% (0) \\
\mdourrow StartLux-Decision-9B, select-then-think, reasoning sets, $\alpha=0.10$, $N=2/4/8$ & Reasoning-route error 22.2 / 18.6 / 15.8\% (20 / 20 / 14) & 6.9--7.8\% (0) \\
\bottomrule
\end{tabular}
\end{adjustbox}\par}
\end{table}

Changes in candidate count $N$, routing rate, generator, or task family may require renewed calibration and threshold certification.

\section{Evaluation Details}\label{sec:D}
This appendix details Decision Index scoring, per-benchmark results, evaluation safeguards, and inference for long option lists and multiple questions.
DecisionPivot is described by~\citet{startlux2026decisionpivot}; its data and scorer are available on \href{https://huggingface.co/datasets/startlux-models/DecisionPivot}{Hugging Face}.

\subsection{Decision Index Evaluation}\label{sec:D.1}
Decision Index 0.2.1 aggregates 38 benchmarks across five weighted domains, with within-domain weights of 1.0 or 1.2.
Individual scores are chance-corrected as $(s-c)/(1-c)$ and clipped to $[0,1]$, where $s$ is accuracy and $c$ is chance accuracy; unanswered requests count as incorrect.
DI 0.3 retains five domains and includes 37 public benchmarks.
Table~\ref{tab:D.1} reports per-benchmark results for the released models and three reference systems.
Repeated evaluations with independently sampled reasoning traces vary by approximately 0.05 index points.

At 27B, AutoThink scores 11.8 points below Instant on MuSR, although invoking reasoning within AutoThink changes accuracy by only +0.5 points.
In a controlled Stage 2 comparison using identical data and the same Stage 1 checkpoint (Figure~\ref{fig:attention-ablation}b), causal and bidirectional models score 50.3 and 64.9 on MuSR, respectively.
Three additional bidirectional seeds score 55.8--58.1, yielding gains of 5.5--14.6 points over the causal model and suggesting that attention direction contributes to the cross-variant difference.
On SATA-Bench, Stage 3 and explicit reasoning account for 4.4 of the 6.7-point cross-variant difference.

\begingroup
\fontsize{7}{8.1}\selectfont\setlength{\tabcolsep}{2pt}\renewcommand{\arraystretch}{1.0}
\begin{longtable}{>{\raggedright\arraybackslash}p{\mdw{0.16}}>{\raggedleft\arraybackslash}p{\mdw{0.08}}>{\raggedleft\arraybackslash}p{\mdw{0.07}}>{\raggedleft\arraybackslash}p{\mdw{0.08}}>{\raggedleft\arraybackslash}p{\mdw{0.07}}>{\raggedleft\arraybackslash}p{\mdw{0.08}}>{\raggedleft\arraybackslash}p{\mdw{0.07}}>{\raggedleft\arraybackslash}p{\mdw{0.075}}>{\raggedleft\arraybackslash}p{\mdw{0.075}}>{\raggedleft\arraybackslash}p{\mdw{0.09}}>{\raggedleft\arraybackslash}p{\mdw{0.07}}>{\raggedleft\arraybackslash}p{\mdw{0.08}}}
\caption{\textbf{Per-benchmark results on the public subset of DI 0.3.} At least one StartLux-Decision model leads on 23 of 37 benchmarks. Scores are chance-corrected skill points (0: chance; 100: perfect accuracy). Benchmarks marked $\bigstar$ have within-domain weight 1.2; others have weight 1.0. Domain weights appear in parentheses. AutoThink uses default predictions (Section~\ref{sec:model-design}). Reference scores are from the public leaderboard on October 7, 2026; best scores are bolded. Most tasks belong to represented training families; Table~\ref{tab:A.2} describes the earlier preview-model mixture.}\label{tab:D.1}\\
\toprule
\textbf{Benchmark} & \multicolumn{2}{c}{\textbf{27B}} & \multicolumn{2}{c}{\textbf{9B}} & \multicolumn{2}{c}{\textbf{4B}} & \textbf{2B} & \textbf{0.8B} & \multicolumn{3}{c}{\textbf{Reference models}} \\
\cmidrule(lr){2-3}\cmidrule(lr){4-5}\cmidrule(lr){6-7}\cmidrule(lr){10-12}
 & \textbf{AutoThink} & \textbf{Instant} & \textbf{AutoThink} & \textbf{Instant} & \textbf{AutoThink} & \textbf{Instant} & \textbf{Instant} & \textbf{Instant} & {\begin{tabular}[t]{@{}r@{}}\textbf{Perplexity}\\\textbf{Decider v1.1}\\\textbf{(27B)}\end{tabular}} & {\begin{tabular}[t]{@{}r@{}}\textbf{Clef}\\\textbf{27B}\end{tabular}} & {\begin{tabular}[t]{@{}r@{}}\textbf{Jev}\\\textbf{(hosted)}\end{tabular}} \\
\midrule
\endfirsthead
\multicolumn{12}{l}{\textit{Table~\thetable{} (continued)}}\\
\toprule
\textbf{Benchmark} & \multicolumn{2}{c}{\textbf{27B}} & \multicolumn{2}{c}{\textbf{9B}} & \multicolumn{2}{c}{\textbf{4B}} & \textbf{2B} & \textbf{0.8B} & \multicolumn{3}{c}{\textbf{Reference models}} \\
\cmidrule(lr){2-3}\cmidrule(lr){4-5}\cmidrule(lr){6-7}\cmidrule(lr){10-12}
 & \textbf{AutoThink} & \textbf{Instant} & \textbf{AutoThink} & \textbf{Instant} & \textbf{AutoThink} & \textbf{Instant} & \textbf{Instant} & \textbf{Instant} & {\begin{tabular}[t]{@{}r@{}}\textbf{Perplexity}\\\textbf{Decider v1.1}\\\textbf{(27B)}\end{tabular}} & {\begin{tabular}[t]{@{}r@{}}\textbf{Clef}\\\textbf{27B}\end{tabular}} & {\begin{tabular}[t]{@{}r@{}}\textbf{Jev}\\\textbf{(hosted)}\end{tabular}} \\
\midrule
\endhead
\bottomrule
\endfoot
\mdgrouprow \textbf{DI 0.3} & \textbf{73.37} & 72.83 & 70.45 & 69.96 & 68.21 & 67.88 & 59.14 & 53.75 & 62.25 & 61.71 & 57.96 \\*
\mdgrouprow \textbf{Knowledge \& Reasoning (25.8\%)} & \textbf{65.40} & 60.65 & 59.42 & 53.27 & 57.48 & 50.49 & 36.03 & 28.45 & 51.96 & 52.80 & 53.87 \\*
\mdbodyrow GPQA Diamond $\bigstar$ & 63.3 & 35.4 & 47.6 & 33.3 & 41.5 & 28.6 & 11.6 & 6.8 & 41.5 & 31.3 & \textbf{71.4} \\
\mdbodyrow GSM8K & \textbf{96.4} & 80.2 & 93.6 & 64.3 & 94.9 & 63.7 & 41.9 & 24.9 & 72.0 & 52.1 & 66.2 \\
\mdbodyrow ChessBench & \textbf{19.4} & 18.6 & 16.6 & 16.7 & 15.1 & 15.1 & 14.1 & 14.2 & 10.3 & 17.8 & 9.8 \\
\mdbodyrow MuSR & 49.7 & 61.5 & 46.1 & 53.3 & 46.3 & 51.2 & 44.0 & 35.1 & 53.5 & \textbf{74.2} & 46.1 \\
\mdbodyrow SATA-Bench & 52.3 & \textbf{59.0} & 53.2 & 57.4 & 53.7 & 57.1 & 53.9 & 50.3 & 27.2 & 32.6 & 25.4 \\
\mdbodyrow CRUXEval & \textbf{93.9} & 78.3 & 74.4 & 50.4 & 80.2 & 43.8 & 5.6 & 0.0 & 77.2 & 78.6 & 57.1 \\
\mdbodyrow CLadder & \textbf{89.2} & 88.7 & 85.5 & 85.1 & 80.0 & 83.6 & 67.0 & 58.7 & 58.6 & 88.2 & 45.3 \\
\mdbodyrow HLE $\bigstar$ & 0.0 & 1.4 & 0.0 & 0.0 & 0.0 & 0.0 & 0.0 & 0.0 & 0.0 & 0.0 & \textbf{4.7} \\
\mdbodyrow MMLU-Pro $\bigstar$ & 80.4 & 69.6 & 72.5 & 60.3 & 68.6 & 55.9 & 37.7 & 25.1 & 67.4 & 61.7 & \textbf{80.5} \\
\mdbodyrow BBH $\bigstar$ & 88.3 & 88.0 & 82.8 & 81.1 & 80.7 & 79.3 & 66.4 & 58.7 & 76.8 & 61.8 & \textbf{89.7} \\
\mdbodyrow WinoGrande $\bigstar$ & 88.0 & \textbf{90.2} & 83.4 & 85.3 & 75.4 & 79.2 & 56.0 & 41.3 & 84.8 & 87.1 & 83.9 \\
\mdgrouprow \textbf{Language Understanding (25.8\%)} & 75.37 & \textbf{75.54} & 71.99 & 73.10 & 69.84 & 71.09 & 62.44 & 55.63 & 67.48 & 61.27 & 59.23 \\*
\mdbodyrow ContractNLI & \textbf{81.3} & 80.6 & 79.6 & 80.4 & 77.9 & 79.0 & 77.2 & 77.4 & 68.7 & 73.0 & 59.1 \\
\mdbodyrow ANLI $\bigstar$ & \textbf{70.5} & \textbf{70.5} & 67.0 & 65.6 & 61.1 & 62.0 & 46.5 & 35.9 & 61.5 & 55.0 & 62.2 \\
\mdbodyrow HellaSwag $\bigstar$ & 97.0 & 97.4 & 95.1 & 95.5 & 91.3 & 93.1 & 84.4 & 71.2 & 91.0 & \textbf{97.6} & 92.7 \\
\mdbodyrow ACOS & 63.4 & \textbf{64.9} & 60.8 & 63.1 & 58.9 & 61.7 & 56.9 & 56.7 & 25.9 & 31.0 & 27.3 \\
\mdbodyrow FinEntity & 93.3 & 93.9 & 93.2 & 93.7 & 93.2 & 93.2 & 90.1 & 87.6 & 93.3 & \textbf{94.2} & 80.8 \\
\mdbodyrow iSarcasmEval & 55.0 & 51.0 & 46.6 & 50.4 & 47.7 & 50.5 & 32.9 & 18.9 & \textbf{62.7} & 31.0 & 36.3 \\
\mdbodyrow VAST & 70.8 & \textbf{71.7} & 68.0 & 68.6 & 68.6 & 67.6 & 66.2 & 63.9 & 65.1 & 39.0 & 46.9 \\
\mdbodyrow NLI4CT & 75.0 & \textbf{75.2} & 67.3 & 70.0 & 64.0 & 63.3 & 45.8 & 36.2 & 70.6 & 66.9 & 69.0 \\
\mdbodyrow RAGTruth & 68.7 & \textbf{71.4} & 66.8 & 67.5 & 63.3 & 66.7 & 60.8 & 53.7 & 65.0 & 57.6 & 51.3 \\
\mdgrouprow \textbf{Retrieval \& Classification (20.0\%)} & 73.97 & \textbf{74.76} & 73.60 & 74.33 & 71.12 & 73.13 & 69.65 & 66.32 & 61.26 & 62.84 & 55.42 \\*
\mdbodyrow BANKING77 $\bigstar$ & 91.7 & 92.3 & 91.7 & 92.9 & 88.6 & 90.7 & 89.7 & 90.0 & 80.3 & \textbf{94.0} & 79.5 \\
\mdbodyrow CLINC150 $\bigstar$ & 94.3 & 93.8 & 94.0 & 94.2 & 92.7 & 93.5 & 90.5 & 90.2 & 91.5 & \textbf{97.4} & 89.2 \\
\mdbodyrow BRIGHT $\bigstar$ & 42.3 & 41.8 & 41.6 & 39.6 & 39.4 & 41.4 & 32.8 & 33.1 & \textbf{44.4} & 40.6 & 40.6 \\
\mdbodyrow Amazon ESCI & 53.9 & \textbf{56.0} & 51.5 & 53.7 & 51.3 & 52.3 & 50.5 & 48.6 & 50.3 & 46.5 & 43.8 \\
\mdbodyrow PhishNChips & 84.2 & 83.5 & \textbf{84.7} & \textbf{84.7} & 82.0 & 81.4 & 77.9 & 63.1 & 46.3 & 58.7 & 25.1 \\
\mdbodyrow HoVer & 76.2 & \textbf{80.4} & 76.8 & 80.1 & 71.3 & 78.2 & 75.7 & 70.0 & 48.2 & 31.2 & 45.7 \\
\mdgrouprow \textbf{Tools \& Automation (18.3\%)} & 86.75 & 87.71 & 85.83 & \textbf{88.14} & 83.98 & 86.44 & 78.62 & 74.61 & 78.88 & 80.71 & 75.09 \\*
\mdbodyrow BFCL $\bigstar$ & 97.4 & 97.3 & 96.8 & 97.0 & 93.0 & 94.9 & 92.8 & 91.5 & 96.2 & \textbf{97.9} & 94.3 \\
\mdbodyrow ToolRet & 63.6 & \textbf{65.2} & 61.3 & 62.7 & 62.5 & 62.9 & 60.6 & 58.6 & 62.6 & 64.3 & 59.9 \\
\mdbodyrow API-Bank $\bigstar$ & 83.8 & 83.4 & 90.0 & 89.2 & 83.4 & 86.8 & 79.1 & 77.7 & 85.2 & \textbf{91.8} & 88.0 \\
\mdbodyrow Home appliances & 98.9 & \textbf{100.0} & 92.0 & 98.9 & 93.2 & 95.5 & 70.5 & 60.2 & 80.7 & 80.7 & 52.3 \\
\mdbodyrow When2Call & 88.6 & \textbf{91.7} & 86.0 & 90.9 & 86.2 & 90.4 & 87.2 & 81.0 & 65.1 & 63.3 & 74.6 \\
\mdgrouprow \textbf{Arts \& Human Taste (10.0\%)} & 63.14 & \textbf{66.20} & 60.55 & 62.98 & 57.07 & 60.11 & 53.65 & 51.00 & 46.96 & 48.88 & 39.08 \\*
\mdbodyrow BPoMP & 95.6 & \textbf{96.5} & 93.6 & 94.8 & 91.1 & 92.1 & 81.9 & 80.6 & 90.5 & 94.1 & 81.8 \\
\mdbodyrow Humicroedit & 35.5 & \textbf{37.7} & 32.1 & 32.8 & 27.8 & 29.2 & 26.2 & 22.8 & 22.3 & 33.1 & 23.7 \\
\mdbodyrow POP909 & 81.2 & \textbf{85.2} & 78.5 & 81.3 & 76.8 & 80.0 & 76.9 & 77.0 & 55.5 & 14.8 & 15.9 \\
\mdbodyrow cfcolor & 43.9 & \textbf{48.1} & 39.1 & 43.9 & 37.6 & 42.7 & 31.9 & 32.8 & 28.1 & 34.0 & 28.8 \\
\mdbodyrow Habermas & 44.9 & 49.2 & 47.0 & 50.0 & 41.4 & 47.0 & 44.4 & 42.2 & 15.8 & \textbf{54.5} & 21.5 \\
\mdbodyrow New Yorker & 77.8 & \textbf{80.6} & 73.0 & 75.1 & 67.8 & 69.7 & 60.7 & 50.5 & 69.5 & 62.8 & 62.6 \\
\end{longtable}
\endgroup

\subsection{GSM8K Benchmark Reconstruction}\label{sec:D.2}
Some distractors in DI 0.2.1's ten-choice GSM8K format provide cues to the correct answer, creating an answer-selection shortcut.
DI 0.3 replaces these distractors with answers from other problems to reduce this artifact.
Earlier checkpoints score 0.7--1.8 points higher on DI 0.2.1 with the original format, whereas the 4B Instant model trained only on reconstructed questions shows no such advantage.

\subsection{Evaluation Reliability and Serving Configuration}\label{sec:D.3}
Valid evaluation runs require at least 99.9\% request completion; unanswered requests count as incorrect.
Prefix caching and chunked prefill are disabled because they can alter decision readouts under the evaluated vLLM 0.19.1 configuration (Table~\ref{tab:D.2}).

\begin{table}[!htbp]
\centering
\vspace{-2mm}
\caption{\textbf{Safeguards for reliable decision evaluation.} Observed effects of vLLM 0.19.1 serving configurations relative to the reference batch runner and the corresponding safeguards. The final row shows the effect of incomplete request coverage.}\label{tab:D.2}
\vspace{-2mm}
{\small\setlength{\tabcolsep}{5pt}\renewcommand{\arraystretch}{1.1}
\noindent\begin{adjustbox}{max width=\linewidth}
\begin{tabular}{>{\raggedright\arraybackslash}p{\mdw{0.26}}>{\raggedright\arraybackslash}p{\mdw{0.29}}>{\raggedright\arraybackslash}p{\mdw{0.27}}>{\raggedright\arraybackslash}p{\mdw{0.18}}}
\toprule
\textbf{Configuration} & \textbf{Observed issue} & \textbf{Measured effect} & \textbf{Safeguard} \\
\midrule
\mdbodyrow Prefix caching enabled & Repeated readouts of the same records differ & 2.6\% of readouts change by more than 0.05 & Disable prefix caching \\
\mdbodyrow Chunked prefill with 8,192 batched tokens (default) & Readouts change when a prompt spans two scheduler steps & 4B with reasoning: 65.01 instead of 66.38 & Process prefill in one step \\
\mdbodyrow Incomplete request coverage & Unanswered requests are scored as incorrect & 61.31 instead of 66.36 & Require at least 99.9\% completion \\
\bottomrule
\end{tabular}
\end{adjustbox}\par}
\end{table}

\subsection{Handling Multiple Questions and Long Option Lists}\label{sec:D.4}
\paragraph{Long Option Lists.}
A single readout supports up to 26 options, each mapped to one letter token.
Longer lists use parallel groups of at most 25 options, advancing the top three from each group to a final readout.
For lists of up to 151 options, this procedure eliminates the correct answer before the final round in 0.2--0.7\% of cases.
On 151-option intent questions, the 4B model achieves 86.7\% accuracy, compared with 95.7\% for 26 options.
Serving such lists requires multiple readouts (Section~\ref{sec:vllm-serving}).

\paragraph{Multiple Questions per Document.}
Questions about a document are evaluated independently, so adding questions does not change existing predictions.
For long documents, the evaluated serving configuration can reuse document representations when the attention pattern permits it.
For a 3,200-token document with 32 questions, this increases measured throughput from 2.8 to 27 decisions per second at 27B.
The gain depends on the serving configuration; bidirectional prompt attention prevents reuse of a question-independent document encoding (Section~\ref{sec:training-attention-effects}).

\mdBibliography
\end{document}